\documentclass[10pt,twocolumn,letterpaper]{article}

\usepackage{cvpr}
\usepackage{times}
\usepackage{epsfig}
\usepackage{graphicx}
\usepackage{amsmath}
\usepackage{amssymb}
\usepackage{longtable}
\usepackage{paralist}
\usepackage{multirow,epstopdf,subfigure}
\usepackage{enumitem}

\usepackage[breaklinks=true,bookmarks=false]{hyperref}

\cvprfinalcopy 

\def\cvprPaperID{****} 

\begin{document}

\title{A$^2$Net: Adjacent Aggregation Networks for Image Raindrop Removal}

\author{Huangxing Lin, \  Xueyang Fu, \  Changxing Jing, \ Xinghao Ding\textsuperscript{*}, \ Yue Huang\\
\normalsize School of Information Science and Engineering, Xiamen University, China\\
{\small \textsuperscript{*}\tt Corresponding author: dxh@xmu.edu.cn}
}

\maketitle

\begin{abstract}
  Existing methods for single images raindrop removal either have poor robustness or suffer from parameter burdens. In this paper, we propose a new Adjacent Aggregation Network (A$^2$Net) with lightweight architectures to remove raindrops from single images. Instead of directly cascading convolutional layers, we design an adjacent aggregation architecture to better fuse features for rich representations generation, which can lead to high quality images reconstruction. To further simplify the learning process, we utilize a problem-specific knowledge to force the network focus on the luminance channel in the YUV color space instead of all RGB channels. By combining adjacent aggregating operation with color space transformation, the proposed A$^2$Net can achieve state-of-the-art performances on raindrop removal with significant parameters reduction. 
\end{abstract}

\section{Introduction}

Severe weather conditions, such as rain \cite{fu2017removing}, haze \cite{he2011single, ren2016single} and snow \cite{wang2017hierarchical}, impact not only human visual perception but also outdoor computer vision systems \cite{liu2018erase}. In rainy days, both dynamic rain streaks and static raindrops adhered to the camera lens can significantly degrade the captured image's quality. The research on rain streaks removal, e.g., \cite{barnum2010analysis, fu2017clearing, fu2017removing, yang} has been well explored. However, these methods cannot be directly used for raindrop removal, since the appearance and physical imaging process of raindrops are completely different from those of rain streaks. Compared with rain streaks, raindrops have more obvious and irregular geometric structures, which makes this problem more challenging. Since most deep networks \cite{he2016deep,huang2017densely} for high-level vision tasks are trained on clean images, their performances are easily affected by adherent raindrops. Thus, designing effective and efficient algorithms for raindrop removal is desirable for a wide range of practical vision applications.

\begin{figure}
	\centering
	\subfigure[Ground truth / SSIM]{\includegraphics[width=1.5in]{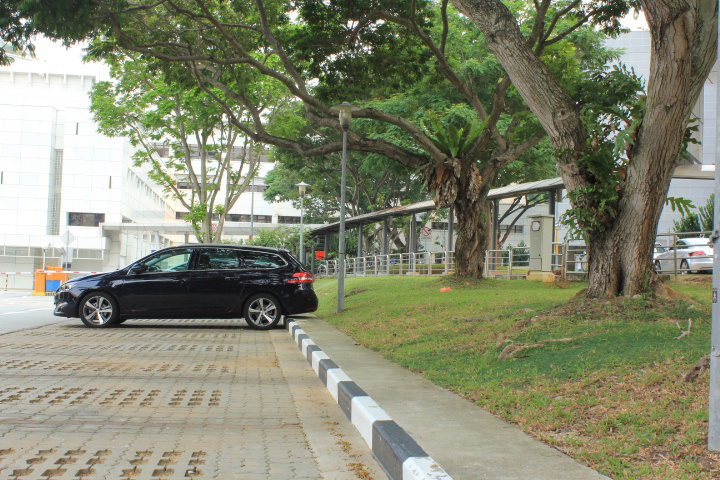}}
	\subfigure[Input / 0.73]{\includegraphics[width=1.5in]{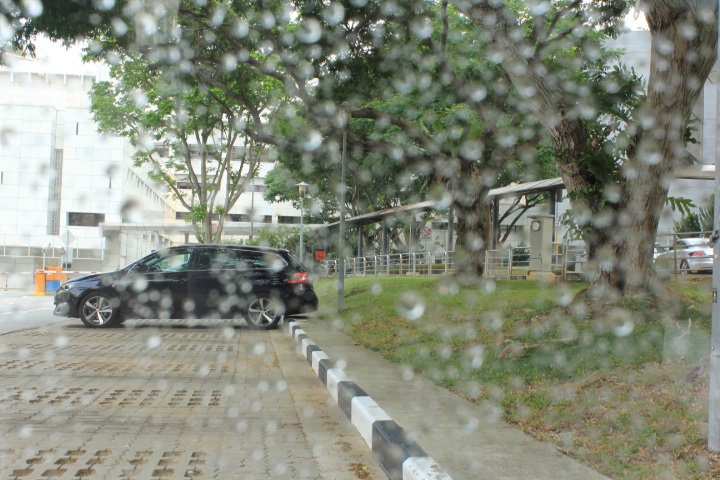}}\\
	\subfigure[AttentiveGAN \cite{qian2018attentive} / 0.85]{\includegraphics[width=1.5in]{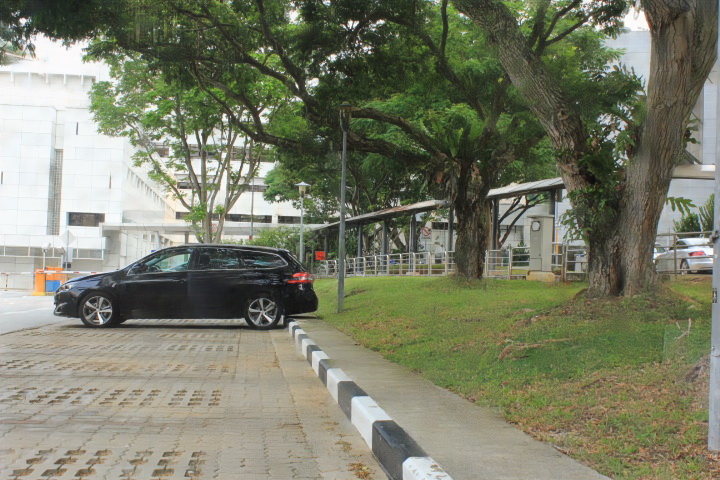}}
	\subfigure[Our / 0.86]{\includegraphics[width=1.5in]{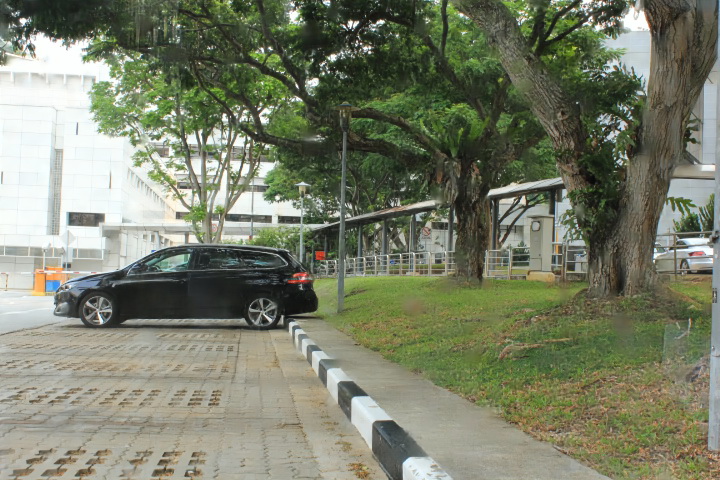}}
	\caption{One example of raindrop removal. Note that the recent AttentiveGAN \cite{qian2018attentive} contains 6.24M parameters, while our A$^2$Net contains only \textbf{0.40M} parameters, which is reduced by 93\%.}
	\label{fig1}
\end{figure}

\begin{figure*}
	\centering
	\includegraphics[width=5.8in]{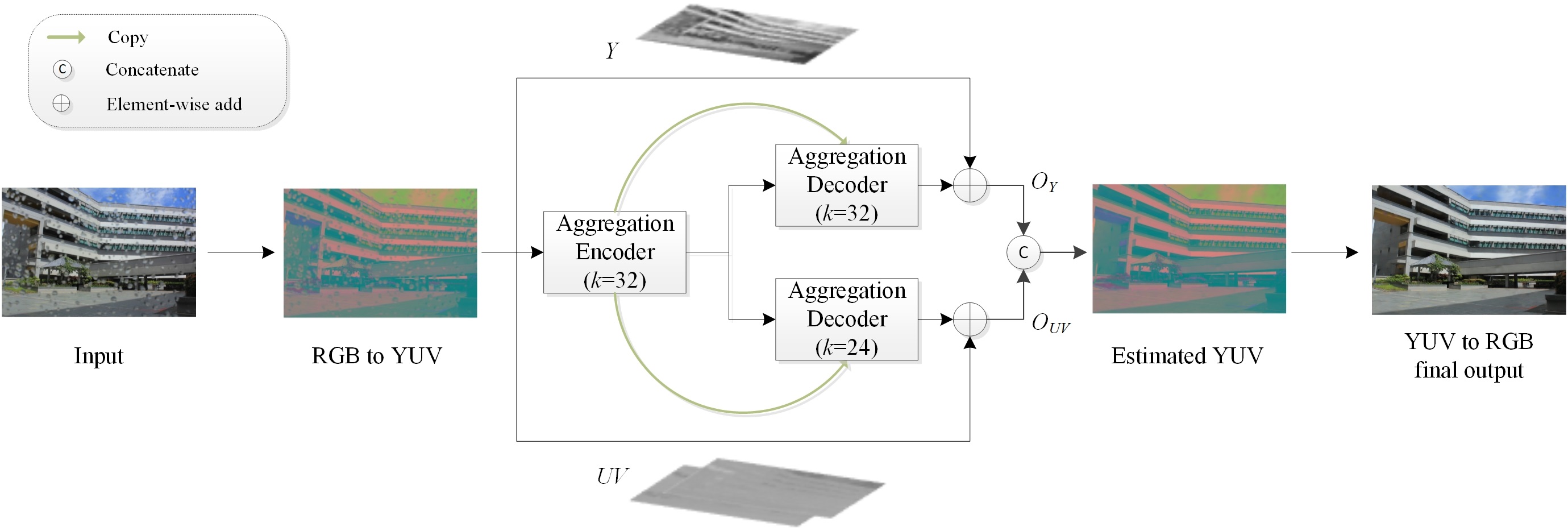}
	\caption{The framework of our A$^2$Net. The input is firstly transformed into the YUV image and all YUV components share one encoder. Then the encoded features of Y and UV are separately decoded as $O_Y$ and $O_{UV}$ to estimate the clean YUV image.  The final output is obtained by transforming the estimated YUV image into RGB image. $k$ denotes the number of feature maps at each convolutional layer.}
	\label{fig2}
\end{figure*}

\subsection{Related works}
To date, many methods have been proposed to handle single image raindrop removal. We briefly review these methods and classify them into two categories: model-based methods and learning-based methods.

\textbf{Model-based methods.} Model-based methods are based on physical imaging process or raindrop geometric appearance to model the distribution of raindrops. In \cite{roser2009video}, the authors attempt to model the shape of adherent raindrops by a sphere section. Furthermore, in \cite{roser2010realistic}, Bessel curves are used to obtain higher modeling accuracy. Since raindrops have various shapes and sizes, the above models can cover only a small portion of raindrop shapes. To simplify the raindrop removal problem, several hardware constraints, e.g., multiple cameras \cite{yamashita2003virtual}, a stereo camera system \cite{tanaka2006removal} and pan-tilt surveillance cameras \cite{yamashita2004removal}, are exploited. However, these methods cannot work with single cameras. In \cite{you2016adherent}, the authors detect raindrops by using motion and intensity temporal derivatives of input videos. Since this method requires consecutive video frames to extract raindrop features, it is not suitable for processing single images.

\textbf{Learning-based methods.} Learning-based methods use large amounts of data to learn and explore the characteristics of raindrops. In \cite{kurihata2005rainy}, the authors detects raindrops on a windshield using raindrop features learned by PCA. When objects in the background are similar to raindrops, PCA cannot effectively extract the characteristics of raindrops and causes false detection. Recently, due to the large amount of available training data and computing resources, deep learning has become the most popular learning-based methods. In \cite{eigen2013restoring}, the authors build a 3 layers network to extract features of static raindrops and dirt spots from synthetic images. While this method works well on small and sparse rain spots, it cannot remove large and dense raindrops. Recently, an AttentiveGAN \cite{qian2018attentive} is proposed to simultaneously detect and remove raindrops. This method employs a recurrent network to detect raindrops and generate corresponding attention maps. These maps are further injected into the following networks to boost the reconstruction performance.

Since raindrops contain various shapes, sizes and appearances, most existing methods cannot simultaneously achieve performance and robustness. The recent AttentiveGAN \cite{qian2018attentive} can well remove raindrops from single images, as shown in Figure \ref{fig1}(c). However, this network contains a relatively large number of parameters and requires complex recurrent operations, which limits its potential value in practical applications with limited computing resources.

\subsection{Our contributions}
\setlength{\parskip}{0pt}
In this paper, we aim to effectively remove the raindrops from single raindrop-degraded images with a lightweight network, as shown in Figure \ref{fig1}(d). To well learn the mapping function between the degraded image and the clean one, not only high-level semantic features but also rich spatial information \cite{yu2017deep} are required. Therefore, to achieve this goal, we design a new feature aggregating operation to better fuse spatial and semantic information across adjacent layers. Furthermore, we find that raindrops are short focal length convex lenses with the function of converging light. The light passing through a raindrop and converge to a point, causing a significant change in the luminance of the raindrop occluded area, while the chrominance will not be affected much. Based on this observation, we force our network to focus on the luminance channel of the YUV color space. Compared with the conventional way of processing all the RGB channels, this color space transformation can significantly simplify the learning process. The contributions of this work are threefold:\\
\vspace{-10pt}
\begin{itemize}
	\item We propose a simple yet effective adjacent aggregation structure which fuses adjacent features for generating more informative representations. By directly deploying this simple aggregation into existing network architectures, the raindrop removal performance can be significantly improved without increasing parameters.
	
	\item We observe a new phenomenon and utilize it for this specific raindrop removal problem. Specifically, instead of directly processing RGB, we intentionally force the network to focus on the luminance (Y channel), but less on the chrominance (UV channels). This divide-and-conquer strategy can significantly simplify the learning process.
	
	\item By combining the aggregating operation with color space transformation, our adjacent aggregation networks (A$^2$Net) can achieve state-of-the-art performance on single images raindrop removal. The A$^2$Net exerts less pressure on system resources (e.g. CPU and memory), which makes it more suitable for practical applications.
	\begin{figure*}
		\centering
		\includegraphics[width=6in]{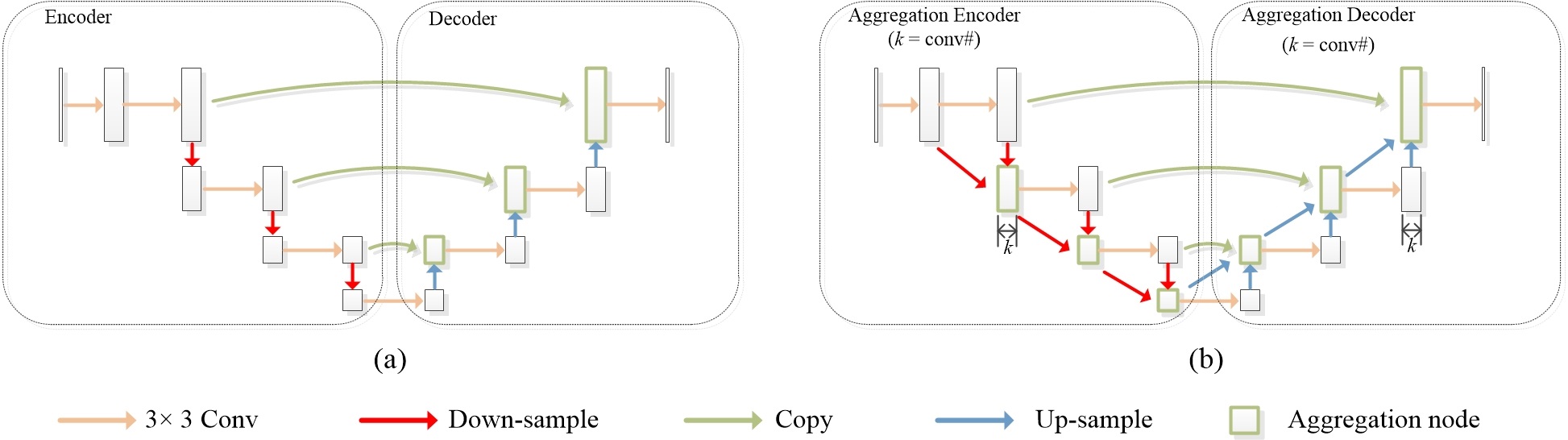}
		\caption{Our adjacent aggregation network. (a) A general encoder-decoder model, (b) Our adjacent aggregation network.}
		\label{A2network}
	\end{figure*}
	
\end{itemize}

\section{Our method}
We show the framework of our proposed A$^2$Net in Figure \ref{fig2}. Instead of designing a complex network architecture, we focus on how to better aggregate adjacent features to generate informative representation for clean image reconstruction. Furthermore, we utilize a problem-specific knowledge to simplify the learning process, i.e., we learn the mapping function in the YUV color space rather than RGB. By this way, the tough single image raindrop removal problem can be well handled with much less parameters.

\subsection{Adjacent aggregation network}
\setlength{\parskip}{0em}
Since raindrops in the image exhibit various shapes, sizes and appearances, it is difficult for a network to identify all raindrops using single-scale features. Naturally, dealing with this problem in a multi-scale fashion is more suitable to capture complex geometric structures of raindrops. Fortunately, the hourglass structured network, e.g., U-Net \cite{ronneberger2015u} and FPN \cite{lin2017feature}, can well extract features at different scales. This classical architecture can benefit the network to identify various raindrops. However, during our experiment, we found that directly using this network cannot generate pleasing results. This is because the hourglass networks are good at extracting high-level semantic features at the cost of detail lost. For the specific raindrop removal problem, the networks not only need to extract strong semantic features to recognize raindrops  but also require rich detail information to reconstruct clean images. The lateral connections, indicated by green lines in Figure \ref{A2network}(a), that fuse low- and high-level features can alleviate this problem to some degree. However, this simple fusion cannot help much to the quality of reconstructed images, since the low-level features are too noisy to provide sufficient semantic guidance \cite{zhang2018exfuse}.

\begin{figure}
	\centering
	\subfigure[2-node for encoder]{\includegraphics[width=1.2in]{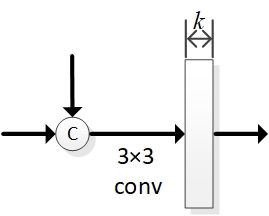}} \hspace{0.2in}
	\subfigure[3-node for decoder]{\includegraphics[width=1.2in]{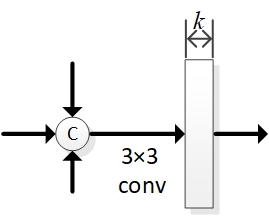}}
	\caption{Illustration of aggregation node architectures. $k$ denotes the number of feature maps.}
	\label{fig4}
\end{figure}

To address the aforementioned problems, we design a new adjacent aggregation network based on the classical encoder-decoder architecture, as shown in Figure \ref{A2network}(b), to better fuse information across layers. Our aggregating operation merges features extracted from adjacent layers and generate new representations with richer detail and semantics. In the aggregation encoder, the inputs of fusion are two adjacent features with same resolutions after down-sampling. In the aggregation decoder, the inputs of fusion are three features with same resolutions, two are adjacent features after up-sampling and one is from the encoder. These features are merged by using same aggregation nodes, as shown in Figure \ref{fig4}. The aggregation operation is defined by
\begin{equation}
\fontsize{10pt}{15.5pt}\selectfont
X=\delta(W*C(X_1,...,X_m)+b),
\end{equation}
where $X_1,...,X_m$ are the inputs of the aggregation node, $C$ denotes the concatenation, $*$ indicates the convolution operation  and $\delta(\cdot)$ is the nonlinear activation. $W$ is a kernel of size $3\times3$ to fuse $X_1,...,X_m$  and the kernel number is a hyper-parameter for each encoder and decoder. $X$ is the fused feature which has the same resolution as $X_1,...,X_m$. In this paper, we set the number of down-sample three times, and the aggregation decoder will correspondingly up-sample three times. Moreover, we set the kernel numbers to the same for all layers in each encoder or decoder based on our experiments.

\begin{figure*}
	\centering
	\subfigure[R channel of $E$]{\includegraphics[width=1.9in]{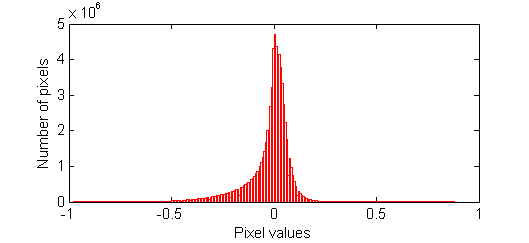}}
	\subfigure[G channel of $E$]{\includegraphics[width=1.9in]{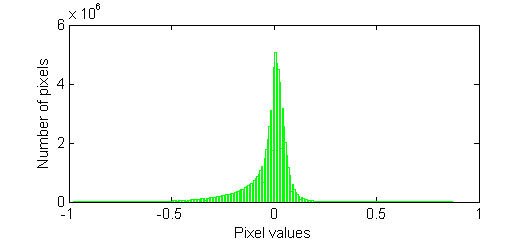}}
	\subfigure[B channel of $E$]{\includegraphics[width=1.9in]{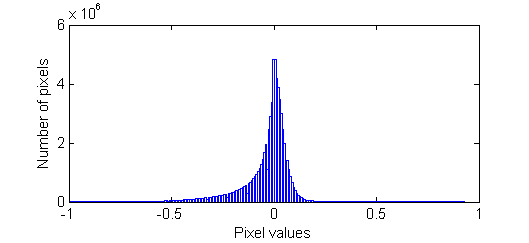}}\\
	\vspace{-1em}
	\subfigure[Y channel of $E_{YUV}$]{\includegraphics[width=1.9in]{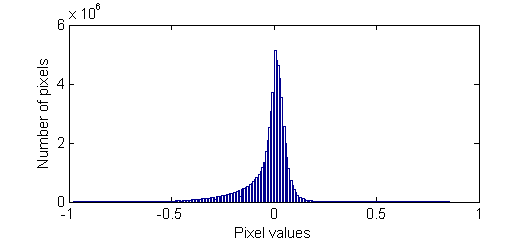}}
	\subfigure[U channel of $E_{YUV}$]{\includegraphics[width=1.9in]{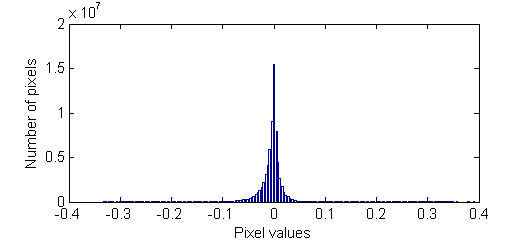}}
	\subfigure[V channel of $E_{YUV}$]{\includegraphics[width=1.9in]{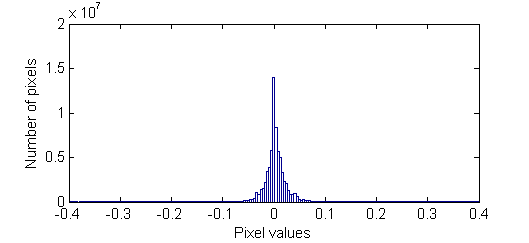}}
	\caption{Residual histogram distributions of 200 pairs of clean and raindrop-degraded images. (a)-(c) represent histograms of residual distributions in R, G, and B channels, respectively. (d)-(f) represent histograms of residual distributions in Y, U, and V channels, respectively.}
	\label{fig5}
\end{figure*}
\begin{figure*}
	\centering
	\includegraphics[width=5in]{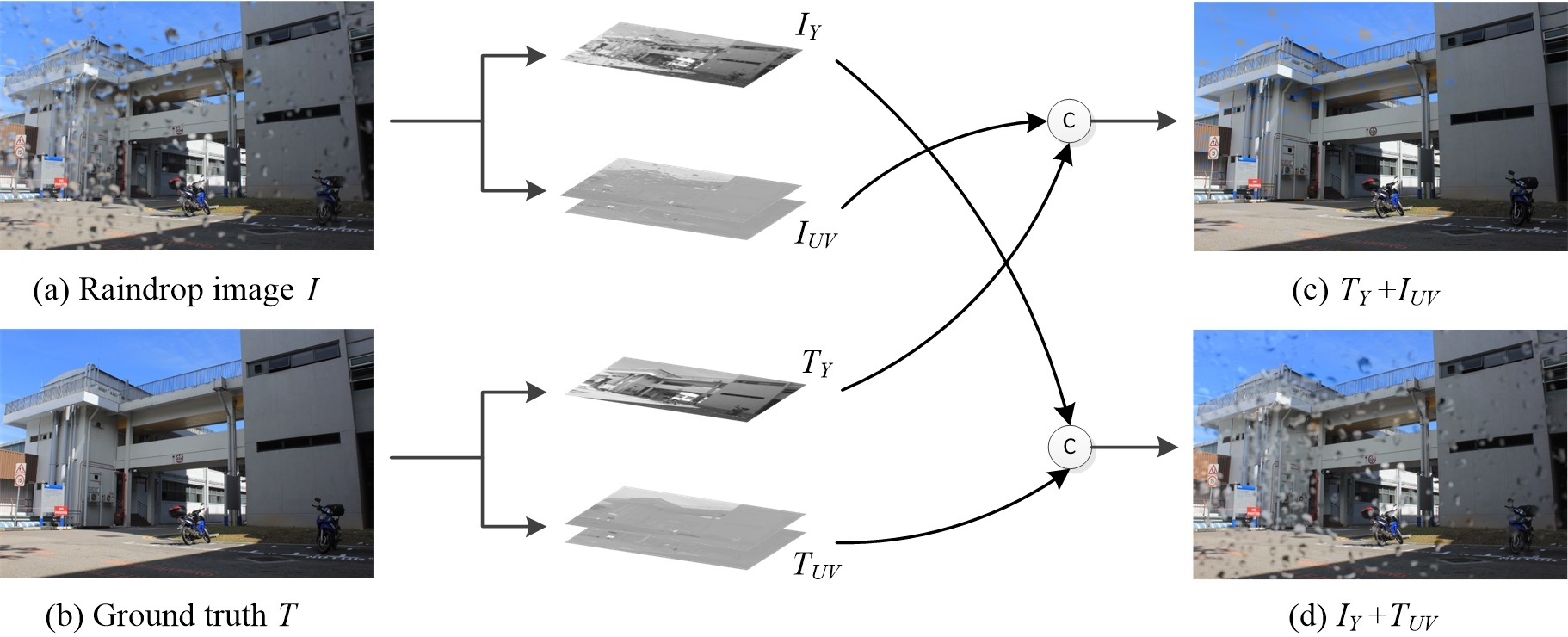}
	\caption{(a) Raindrop image (input) $I$.  (b) Ground truth $T$. (c) An RGB image converted from a concatenation of the Y channel of $T$ and the UV channels of $I$. (d) An RGB image converted from a concatenation of the Y channel of $I$ and the UV channels of $T$.}
	\label{fig6}
\end{figure*}

\subsection{Color space conversion}
To further improve the raindrop removal results, we investigate the way raindrops degenerate images  and observe an interesting phenomenon: the adherent raindrops are short focal length convex lenses with the function of converging light. The rays of reflected light from a wider environment will converge to a point after passing through a raindrop. According to Grassmann's laws \cite{malacara2011color}, when two lights of different color are mixed, they give rise to a color which lies in between the two, while the total luminance of any mixture light is the sum of each light’s luminance. Therefore, there should be obvious changes in the luminance of the raindrop occluded area in the image, while the chrominance will not be affected much. This observation inspires us that if we can remove raindrops from the luminance channel of the image, the quality of the raindrop-degraded image should be greatly enhanced, like shown in Figure \ref{fig6}(c).

Computer vision tasks are dominated by the RGB color space, which seems to be the most intuitive choice. We hereinafter denote the input raindrop-degraded RGB image as $I$ and the corresponding ground truth as $T$. Then the residual $E$ can be represented as
\begin{equation}
E=T-I,
\end{equation}

However, RGB color space does not separate the luminance from the chrominance components \cite{al2014skin}, and pixel values in RGB color space are highly correlated. This makes it a poor choice for luminance and chrominance analysis. Therefore, we convert $I$ from the RGB color space into the YUV color space, where the luminance (Y channel) and chrominance (UV channels) are separated. The residual in YUV color model can be expressed as

\begin{equation}
E_{YUV}=T_{YUV}-I_{YUV},
\end{equation}
\begin{figure*}
	\centering
	\subfigure[Adjacent features 1]{\includegraphics[width=2.2in,trim=50 180 50 150,clip]{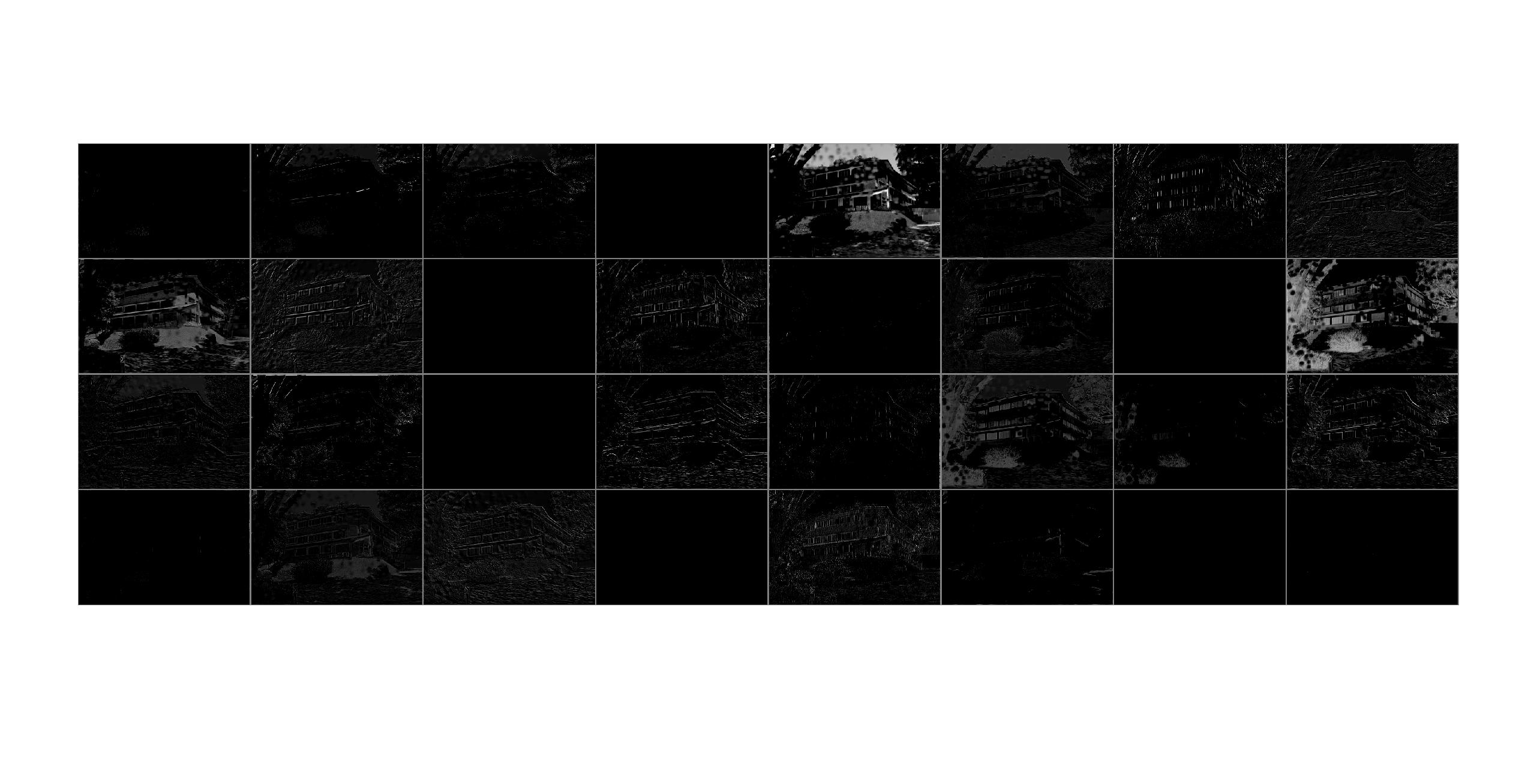}}
	\subfigure[Adjacent features 2]{\includegraphics[width=2.2in,trim=50 180 50 150,clip]{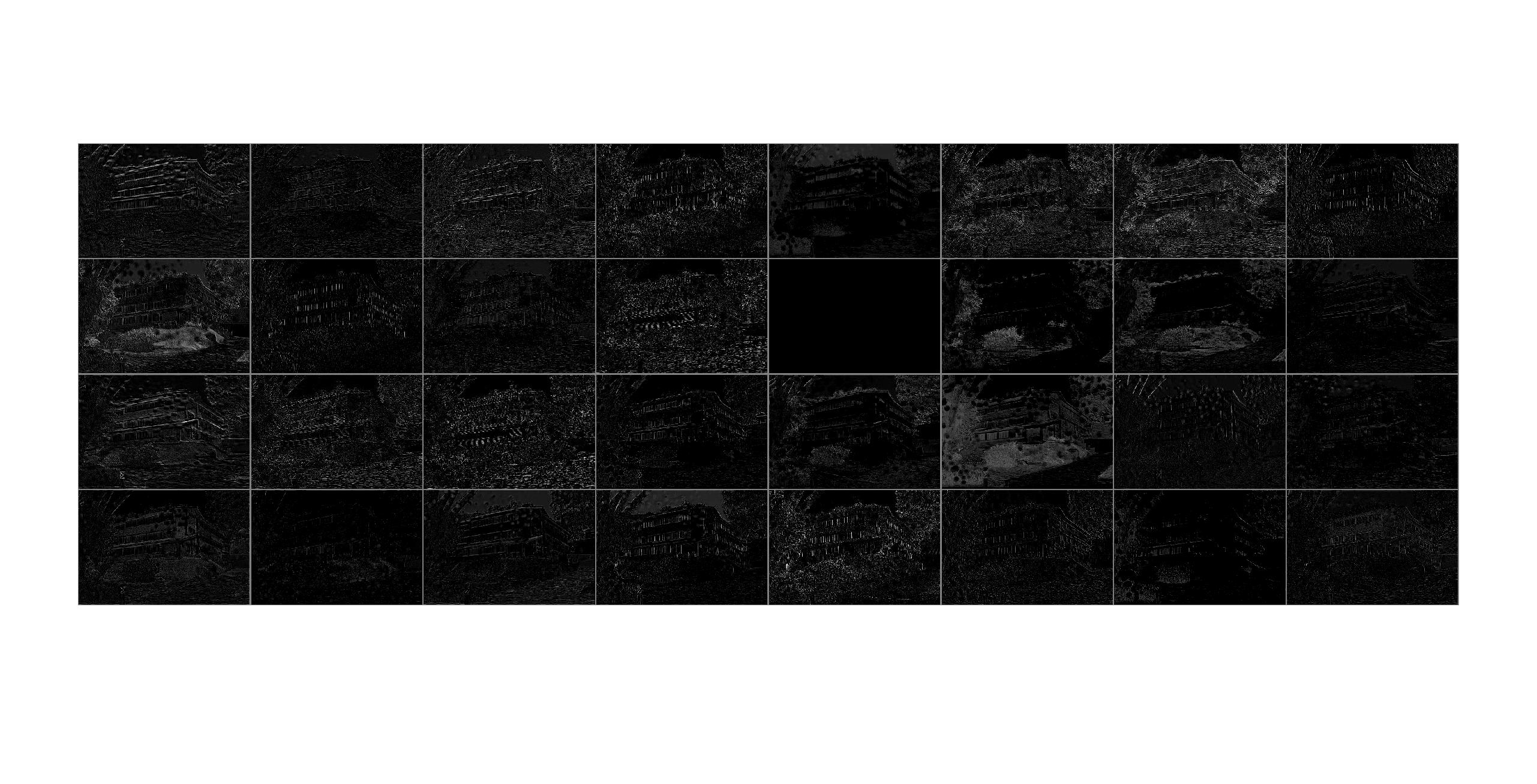}}
	\subfigure[Fusing (a) and (b)]{\includegraphics[width=2.2in,trim=50 180 50 150,clip]{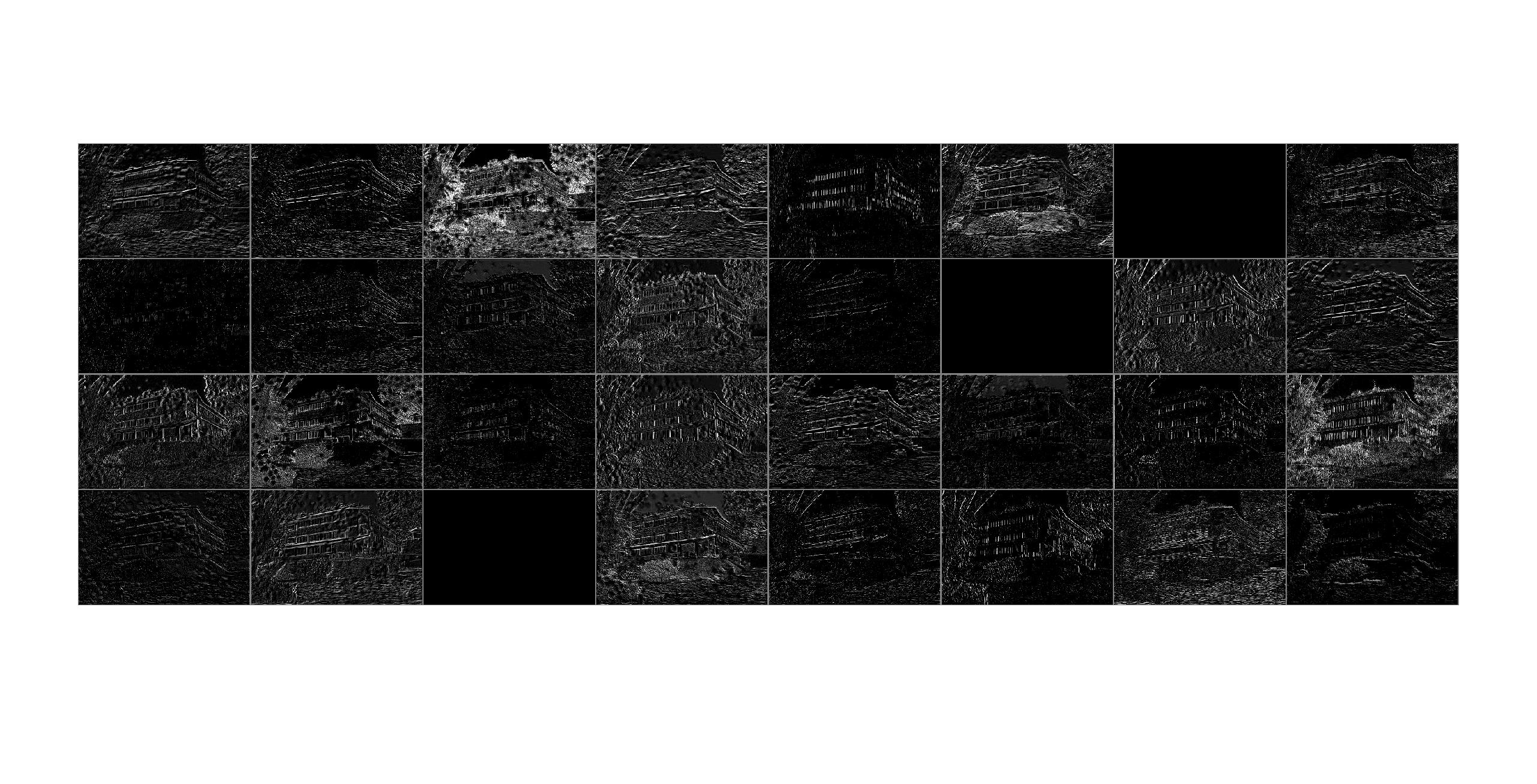}}
	\caption{The aggregation result of the first aggregation node in the encoder.}
	\label{fig7}
\end{figure*}
\vspace{-0.65in}
\begin{figure*}
	\centering
	\subfigure[Adjacent features 1]{\includegraphics[width=1.65in,trim=50 180 50 160,clip]{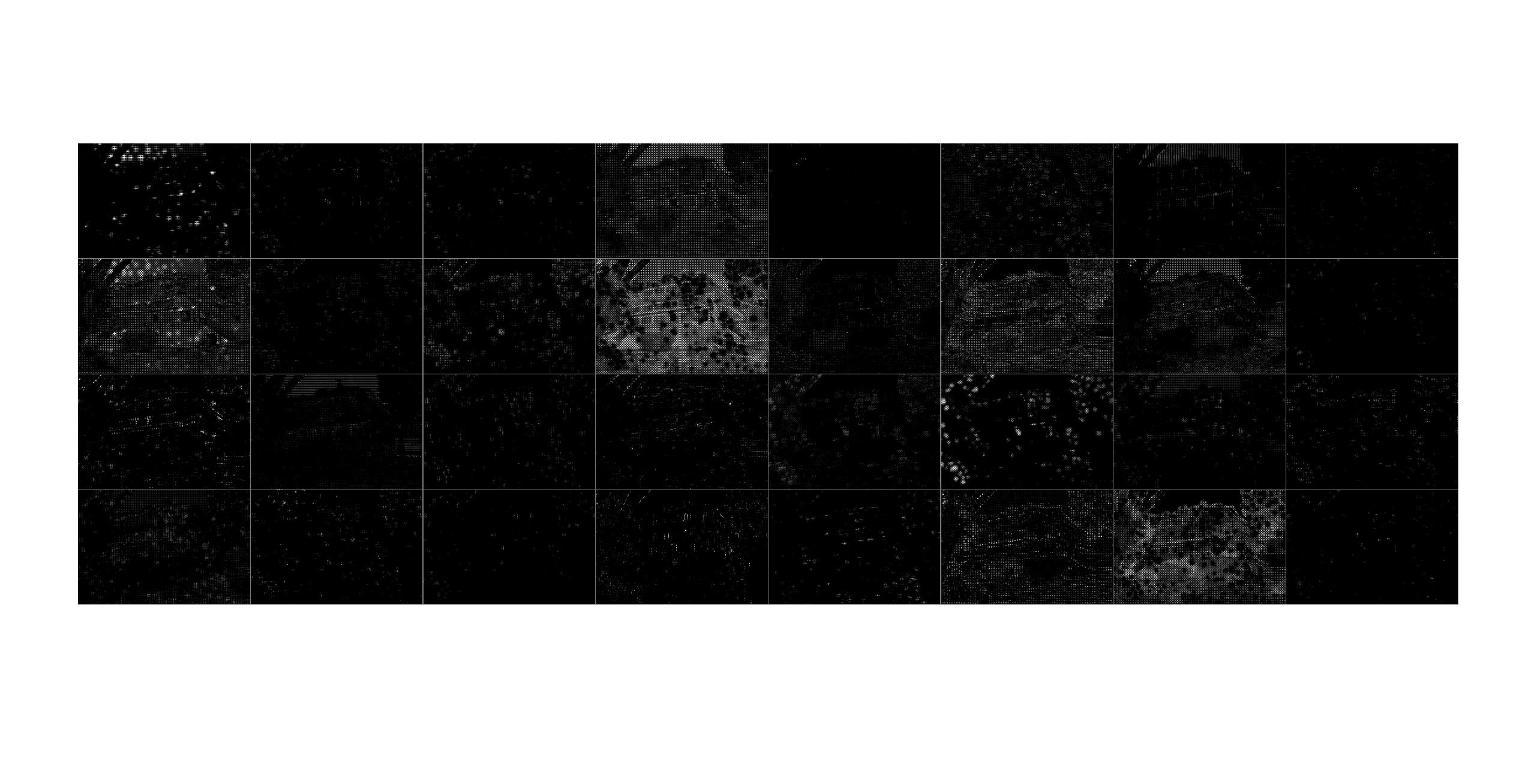}}
	\subfigure[Adjacent features 2]{\includegraphics[width=1.65in,trim=50 180 50 160,clip]{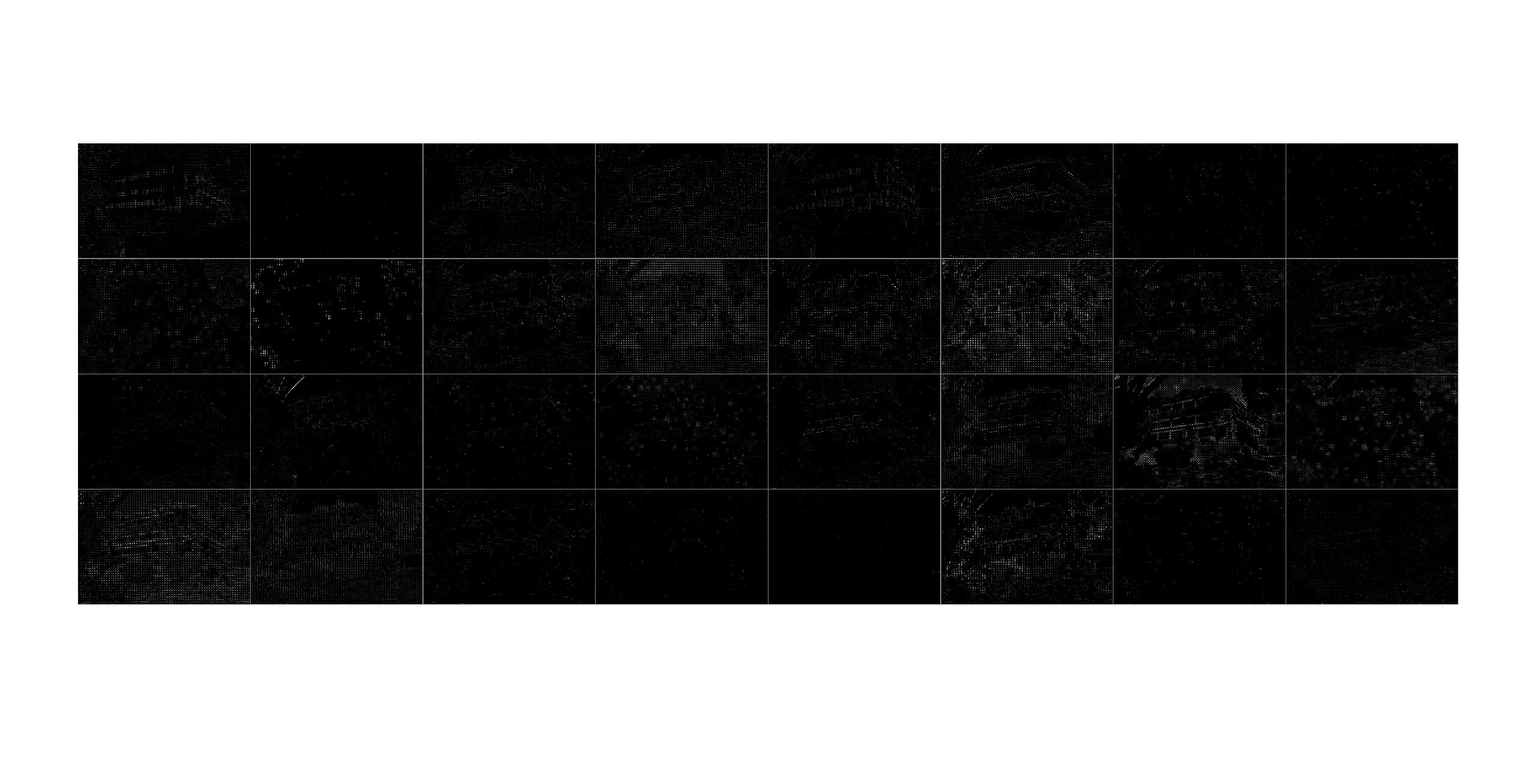}}
	\subfigure[Low-level features]{\includegraphics[width=1.65in,trim=50 180 50 160,clip]{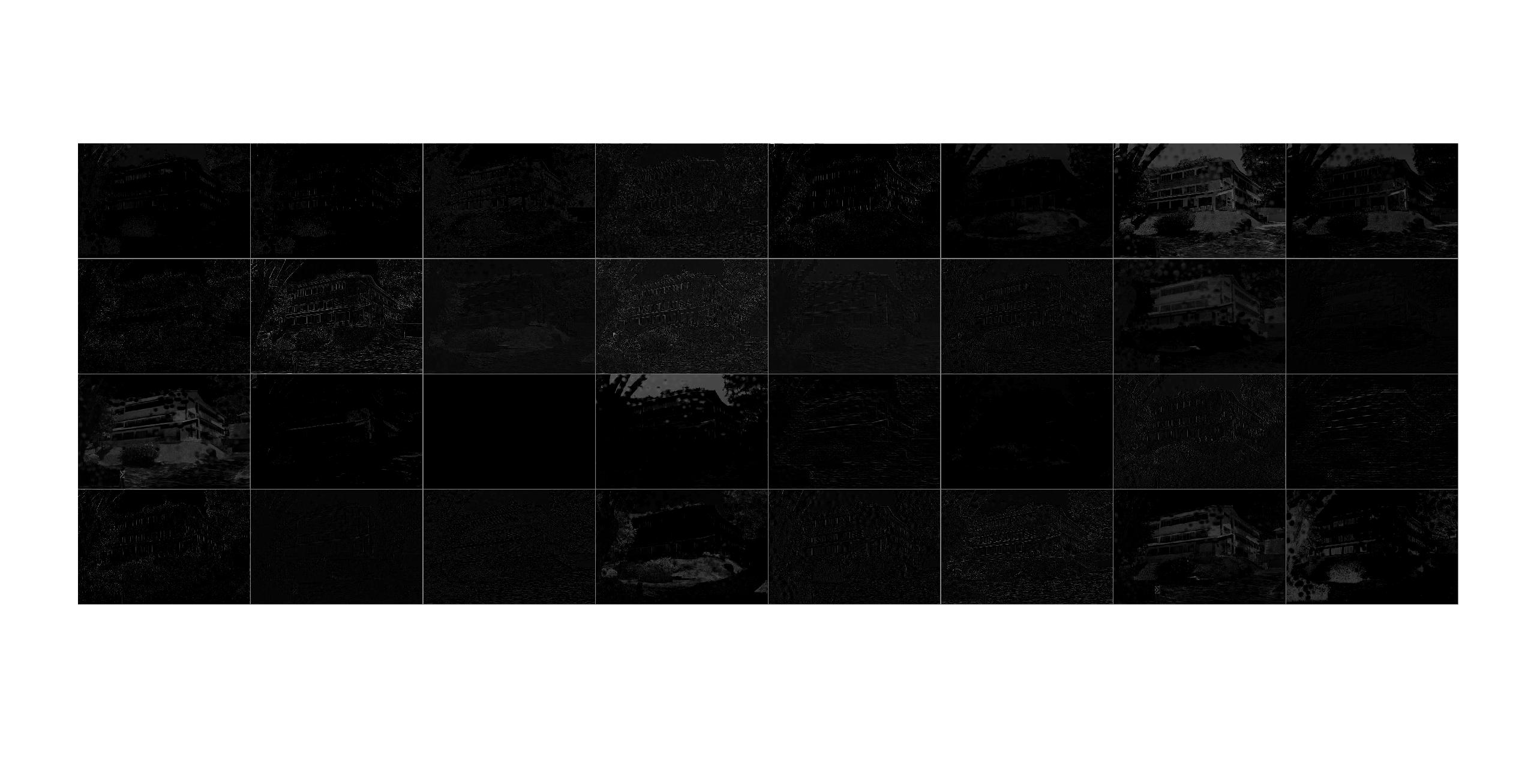}}
	\subfigure[Fusing (a), (b) and (c)]{\includegraphics[width=1.65in,trim=50 160 50 160,clip]{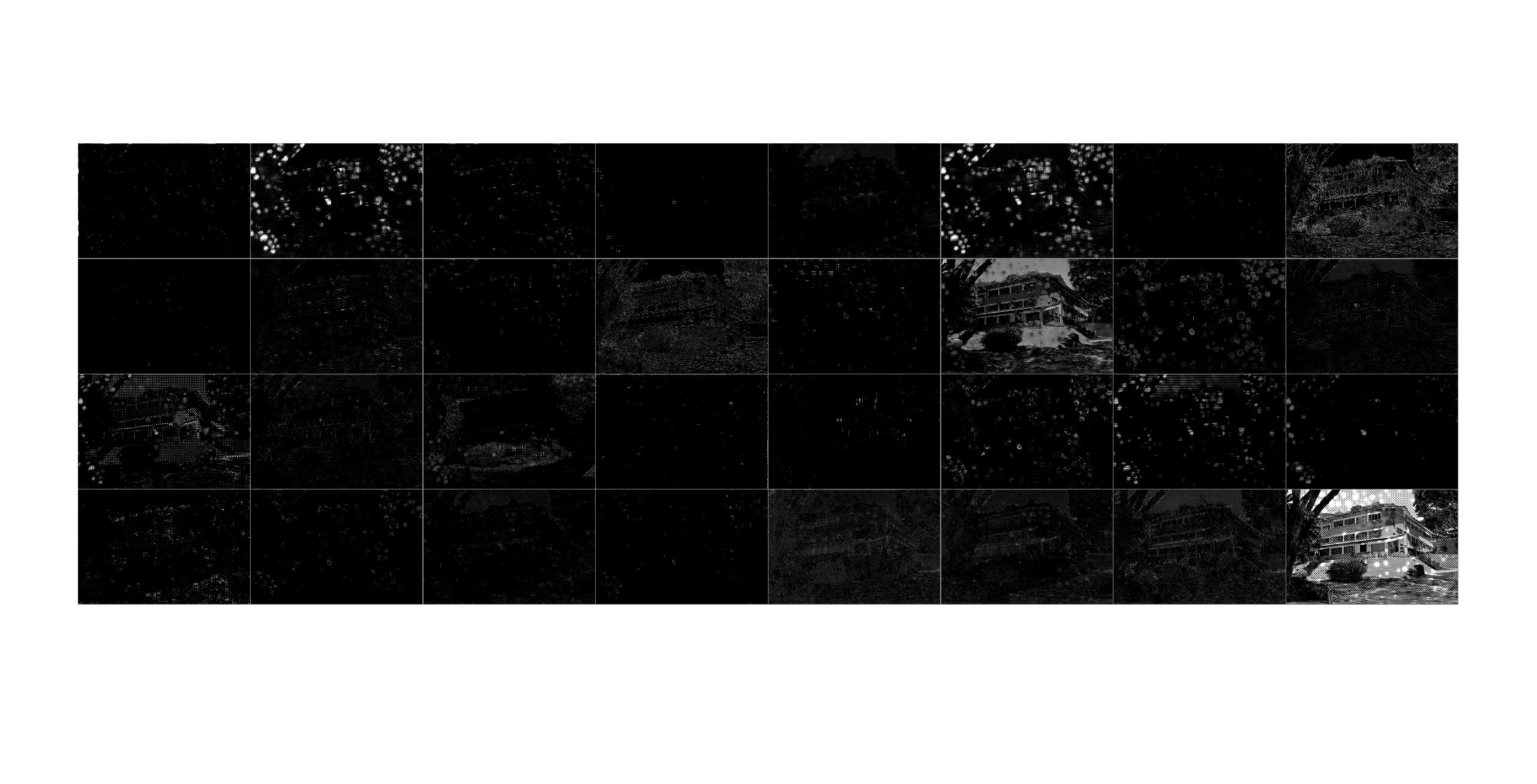}}
	\caption{The aggregation result of the last aggregation node in the decoder for Y channel. Please zoom in for a better visualization.}
	\label{map}
\end{figure*}
\begin{flushleft} \noindent where the subscript $‘YUV’$ denotes the YUV color model.\end{flushleft}

To verify the effectiveness of our color space conversion strategy is, we randomly select 200 pairs of raindrop-degraded images and the corresponding raindrop-free images from the raindrop dataset \cite{qian2018attentive}. Figure \ref{fig5} shows the residual histogram distributions over all 200 pairs of images. Since channels in RGB space are strongly coupled, raindrops have almost the same effect on each channel in RGB space, as shown in Figures \ref{fig5}(a) to (c). While in the YUV space, the residual components in UV channels have significant range reduction in pixel value compared with Y channel, as shown in Figures \ref{fig5}(d) to (f). This implies that the impact of raindrops on the image is mainly concentrated in the luminance channel. Furthermore, using a simple channel swap operation, we observe that raindrops are rare in the UV channels (as shown in Figure \ref{fig6}(c)), while most of the raindrops remain in the Y channel (as shown in Figure \ref{fig6}(d)).

Given the above, the YUV space should be a better choice for raindrop removal than RGB space. In the YUV space, luminance (Y) and chrominance (UV) information can be processed separately, so we can divide the single image-to-image mapping problem into two easier sub-problems. In addition, since most raindrops exhibit in the luminance channel, we can intentionally force the network to focus on the Y channel to simplify the learning process.

\subsection{Proposed A$^2$Net}
Combining with the adjacent aggregation and color space conversion strategy, we propose the A$^2$Net as shown in Figure \ref{fig2}. Our A$^2$Net contains one aggregation encoder and two aggregation decoders. One decoder is for Y channel and the other is for UV channels. The lateral connections are used to connect the encoder and decoders for information propogation. In this way, the network can achieve better raindrop removal results and parameters efficiency than directly learning in RGB space or YUV space, as shown in Figure \ref{fig12}. At the same time, we find that Y and UV channels can share one encoder to guide each other for better raindrop removal. Our network adopts global residual learning with the long short-cut to ease the learning process \cite{fu2017removing}. The output components are represented as $O_Y$ and $O_{UV}$, which are concatenated to generate the estimated clean YUV image. Finally, we convert the YUV image to the RGB image. We set the kernel number of the decoder in Y channel larger than that in the UV channels, so that our network can mainly concentrate on the Y channel. In the final A$^2$Net, the kernel numbers are set to 32, 32 and 24, as shown in Figure \ref{fig2}.

To verify the effectiveness of the proposed aggregation operation, we show the learned feature maps in Figure \ref{fig7} and Figure \ref{map}. Figure \ref{fig7} shows the aggregation result of the first aggregation node in the encoder. Figures \ref{fig7}(a) and (b) are the adjacent feature maps. As can be seen, the two feature maps contain many invalid information, while the fused features are more informative as shown in Figure \ref{fig7}(c). The fused features retain more details while gaining stronger semantics, which can bring benefits for reconstructing images. Figure \ref{map} shows the aggregation result of the last aggregation node in the decoder for Y channel. The fused features (as shown in Figure \ref{map}(d)) can better capture the raindrops, which makes it easier for the decoder to estimate the residual component in Y channel. We argue that the performance of raindrop removal can be significantly enhanced by effectively aggregating adjacent features.


\subsection{Loss function}
The mean squared error (MSE) is widely used to train a network. However, raindrops are blend with objects and background scene, it is hard to distinguish between raindrops and objects by simply minimizing MSE loss. A model with only MSE loss tends to result in a blurred reconstruction. Therefore, for Y and UV channels we adopt MSE + SSIM loss \cite{wang2004image} which can preserve global structure better as well as keeping per-pixel similarity. We minimize the combination of those loss functions to train the encoder and decoders. The loss functions for Y and UV channels are defined by:
\begin{equation}
\fontsize{8.5pt}{12.25pt}\selectfont
\label{eq.4}
L_Y=\frac{1}{M}\sum_{i=1}^{M}(L_{MSE}(O_Y^i,T_Y^i)+L_{SSIM}(O_Y^i,T_Y^i)),
\end{equation}
\begin{equation}
\fontsize{8.5pt}{12.25pt}\selectfont
\label{eq.5}
L_{UV}=\frac{1}{M}\sum_{i=1}^{M}(L_{MSE}(O_{UV}^i,T_{UV}^i)+L_{SSIM}(O_{UV}^i,T_{UV}^i)),
\end{equation}

\noindent where $M$ is the number of training data, $T_Y$ and $T_{UV}$ denote the components of ground truth, $L_{MSE}$ and $L_{SSIM}$ are the MSE and SSIM loss, respectively. The overall loss function can be written as:
\begin{equation}
\fontsize{9pt}{13.5pt}\selectfont
\label{eq.6}
L=L_Y+\alpha \cdot L_{UV},
\end{equation}
where $\alpha$ is the parameter to balance the two losses. We set $\alpha=0.6$ based on cross validation, which forces the network to focus on the Y channel.

\begin{figure*}
	\centering
	\includegraphics[width=1.1in]{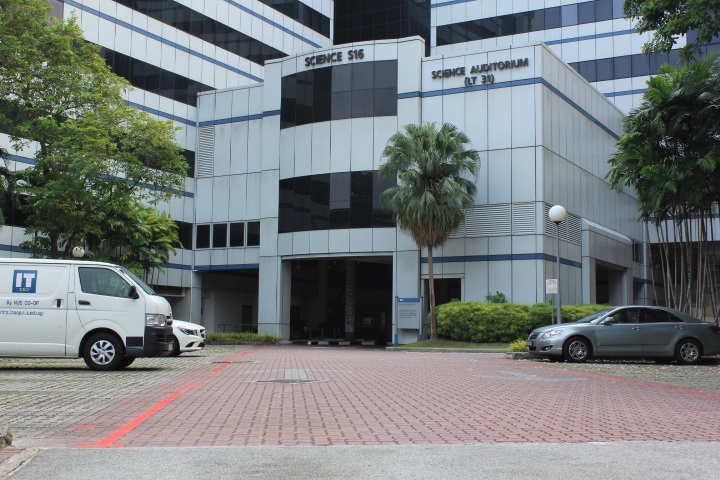}
	\includegraphics[width=1.1in]{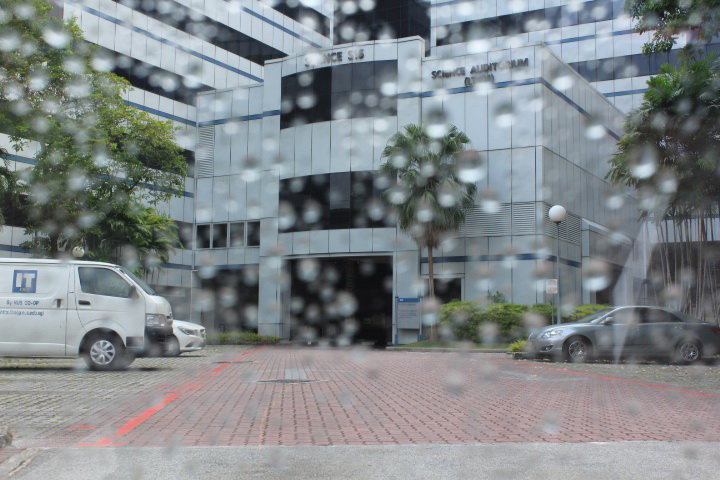}
	\includegraphics[width=1.1in]{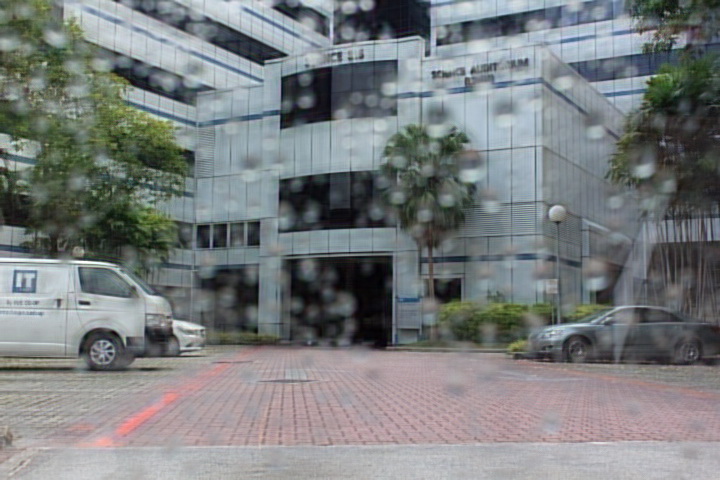}
	\includegraphics[width=1.1in]{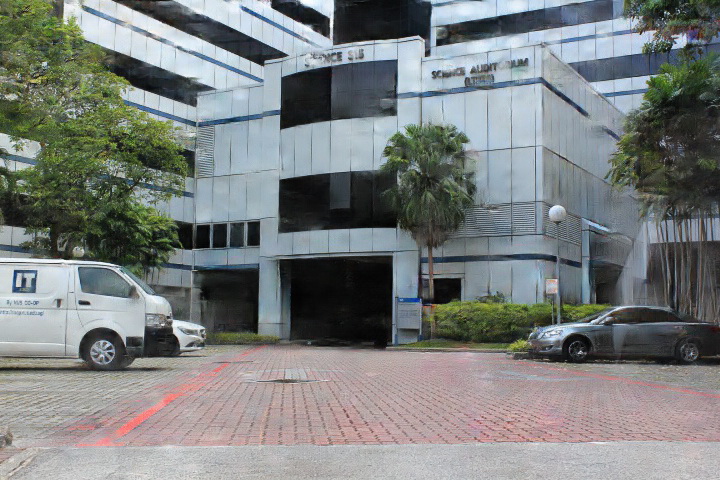}
	\includegraphics[width=1.1in]{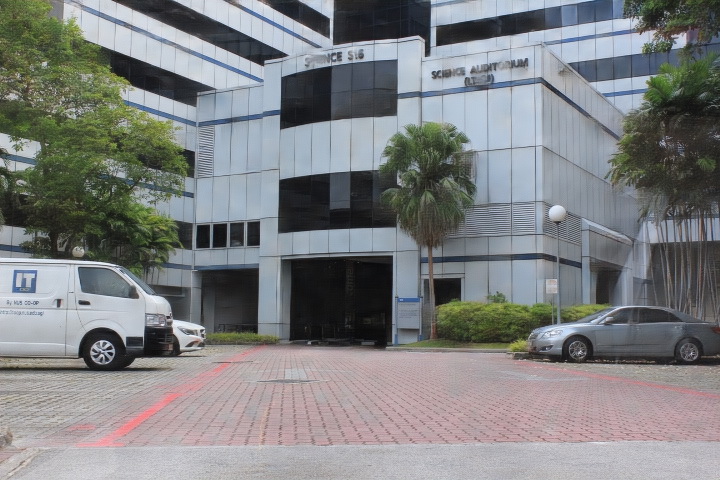}
	\includegraphics[width=1.1in]{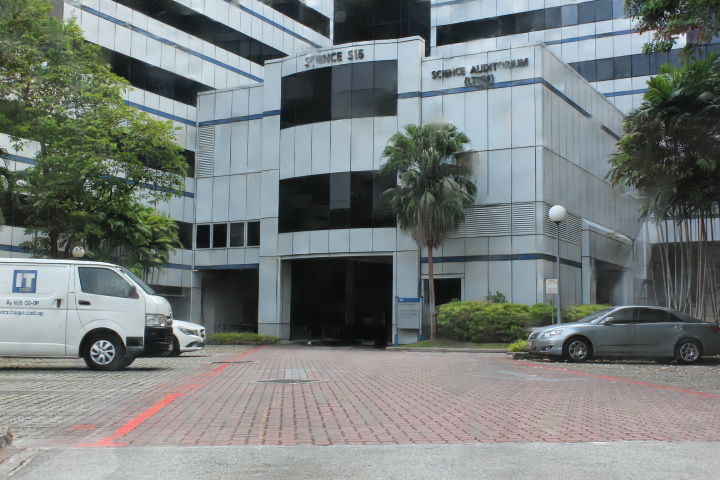}\\
	\includegraphics[width=1.1in]{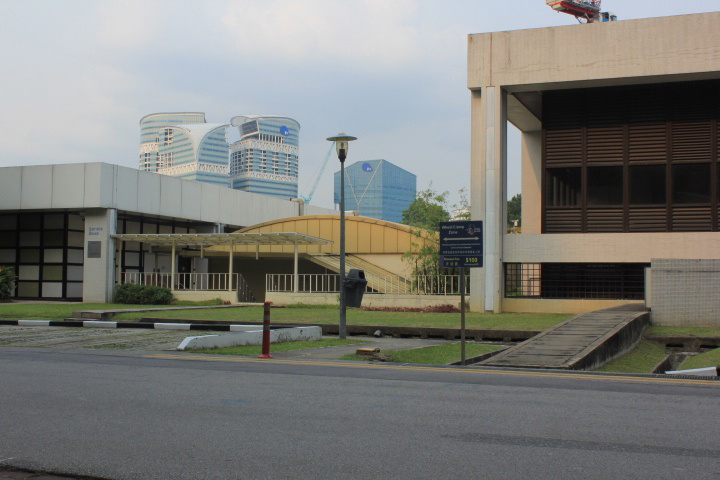}
	\includegraphics[width=1.1in]{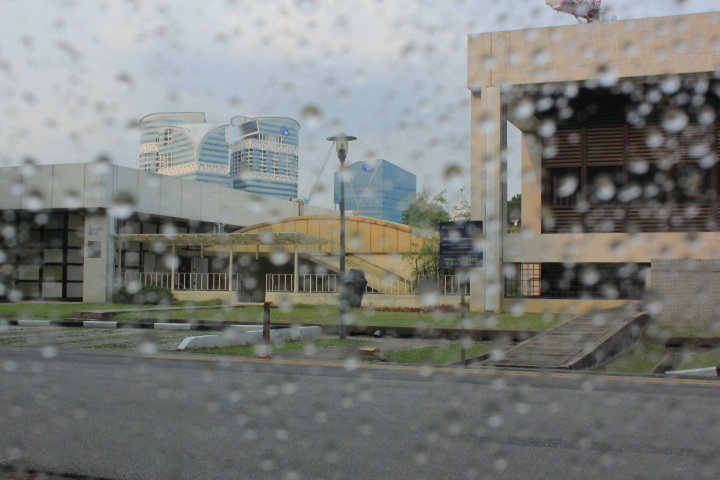}
	\includegraphics[width=1.1in]{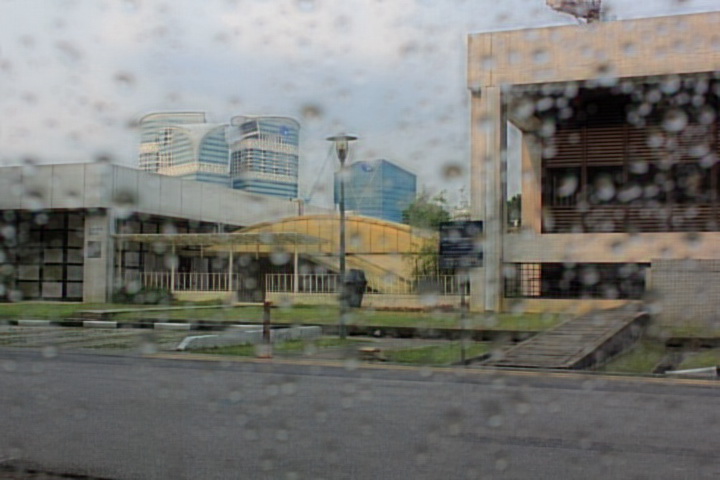}
	\includegraphics[width=1.1in]{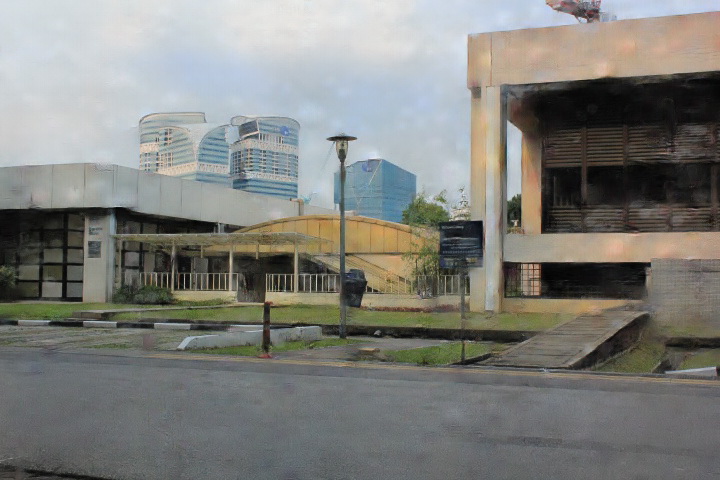}
	\includegraphics[width=1.1in]{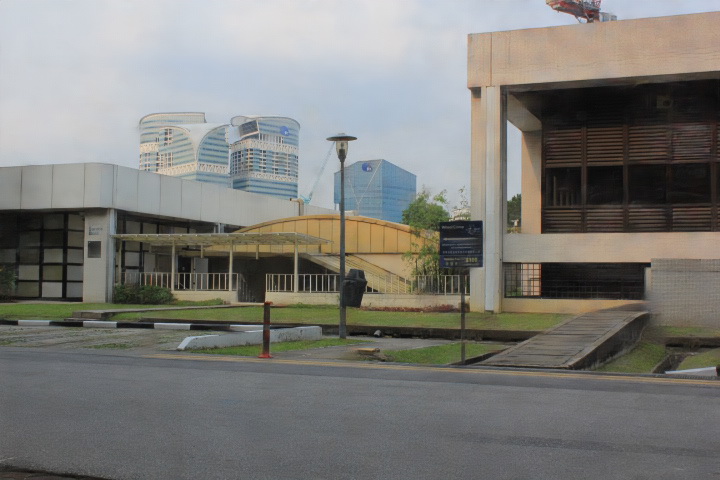}
	\includegraphics[width=1.1in]{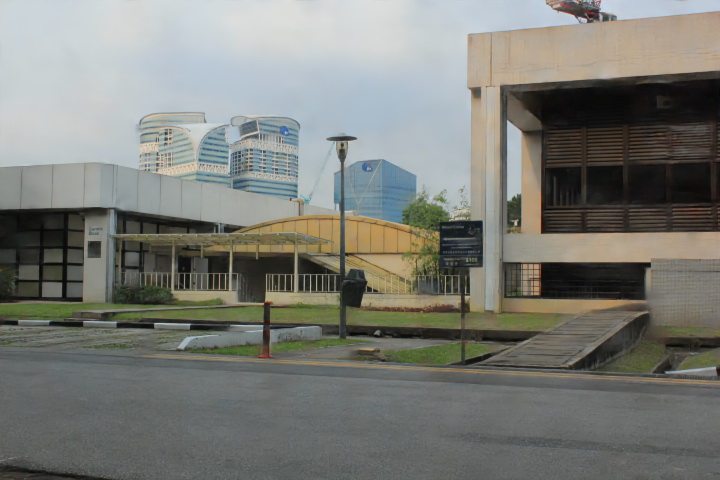}\\  \vspace{-0.065in}
	\subfigure[Ground truth]{\includegraphics[width=1.1in]{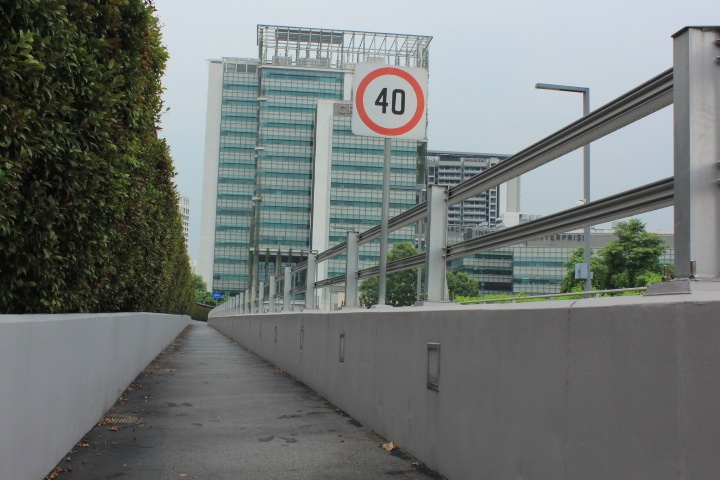}}
	\subfigure[Input]{\includegraphics[width=1.1in]{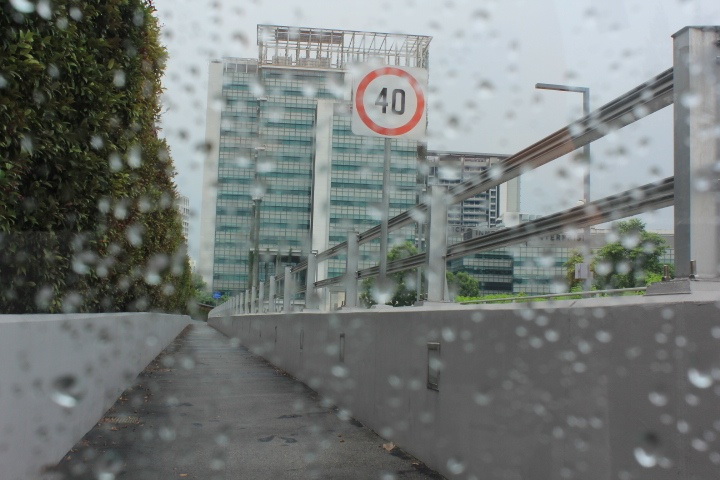}}
	\subfigure[Eigen \cite{eigen2013restoring}]{\includegraphics[width=1.1in]{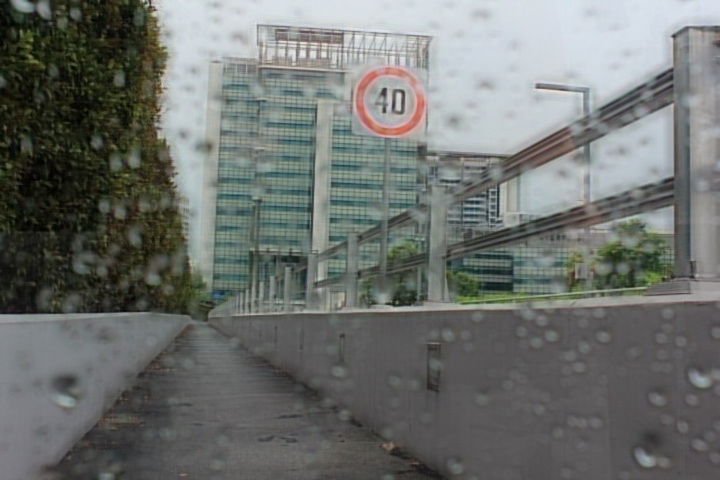}}
	\subfigure[Pix2pix \cite{isola2017image}]{\includegraphics[width=1.1in]{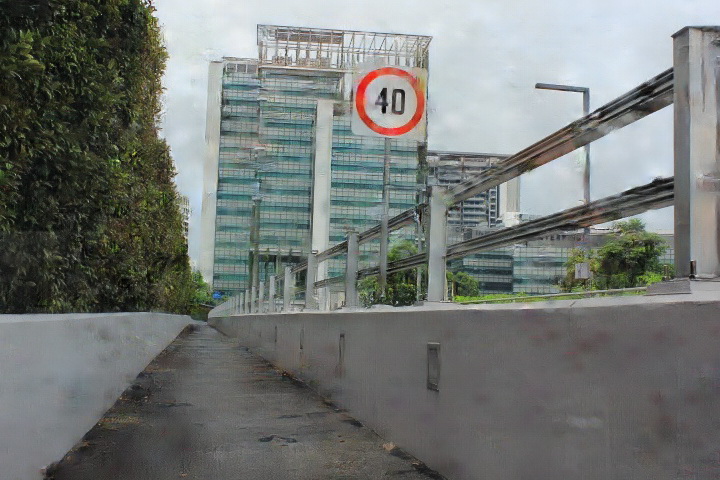}}
	\subfigure[AttentiveGAN \cite{qian2018attentive}]{\includegraphics[width=1.1in]{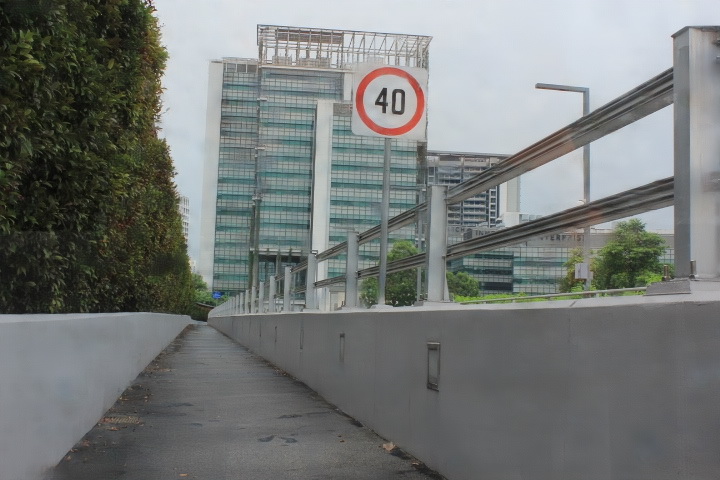}}
	\subfigure[Our A$^2$Net]{\includegraphics[width=1.1in]{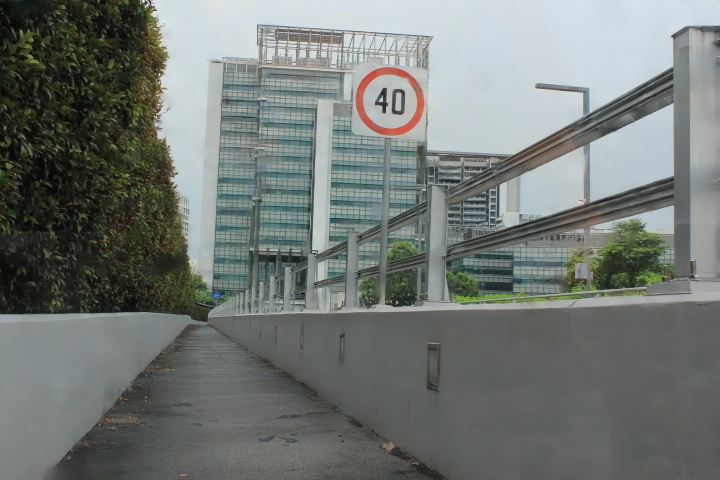}}
	\caption{Visual comparisons on raindrop dataset provided by \cite{qian2018attentive}.}
	\label{fig9}
\end{figure*}
\begin{table*}
	\caption{Quantitative evaluation results. Parameters reduction is shown in parentheses.}
	\centering
	\begin{tabular}{|c|c|c|c|c|c|c|c|c|c|c|c|}
		\hline
		\multicolumn{2}{|c|}{} & \multicolumn{2}{|c|}{Eigen \cite{eigen2013restoring}} & \multicolumn{2}{|c|}{Pix2pix \cite{isola2017image}} & \multicolumn{2}{|c|}{AttentiveGAN \cite{qian2018attentive}} & \multicolumn{2}{|c|}{Our A$^2$Net}\\
		\hline
		\multicolumn{2}{|c|}{}      &SSIM & PSNR&SSIM & PSNR&\ \ SSIM \ \ & PSNR&SSIM & PSNR \\
		\hline
		\multicolumn{2}{|c|}{testA} &0.80 & 23.90&0.84 & 27.13&0.92 & \textbf{31.60}&\textbf{0.93} & 30.79 \\
		\hline
		\multicolumn{2}{|c|}{testB} &0.71 & 21.83&0.76 & 23.34&0.81 & 24.69&\textbf{0.83} & \textbf{25.50} \\
		\hline
		\multicolumn{2}{|c|}{Parameters \#} & \multicolumn{2}{|c|}{0.75M ($\downarrow$ 46\%)} & \multicolumn{2}{|c|}{41.83M ($\downarrow$ 99\%)} & \multicolumn{2}{|c|}{6.24M  ($\downarrow$ 93\%)} & \multicolumn{2}{|c|}{\textbf{0.40M}}\\
		\hline
	\end{tabular}
	\label{table1}
\end{table*}

\subsection{Parameter setting}
As discussed, the proposed A$^2$Net consists of one aggregation encoder and two aggregation decoders. Each encoder and decoder includes several convolutional kernels of size $3\times3$. In the encoder, the down-sampling operations are performed by $4\times4$ convolution with stride $2$. In the decoders, we use $2\times2$ deconvolution with stride 2 to up-sample the low-resolution features. The activation functions for all convolutional and deconvolution layers are $ReLU$ \cite{krizhevsky2012imagenet}, while the activation function of the last layer is $tanh$.

\begin{figure*}
	\centering
	\includegraphics[width=1.2in]{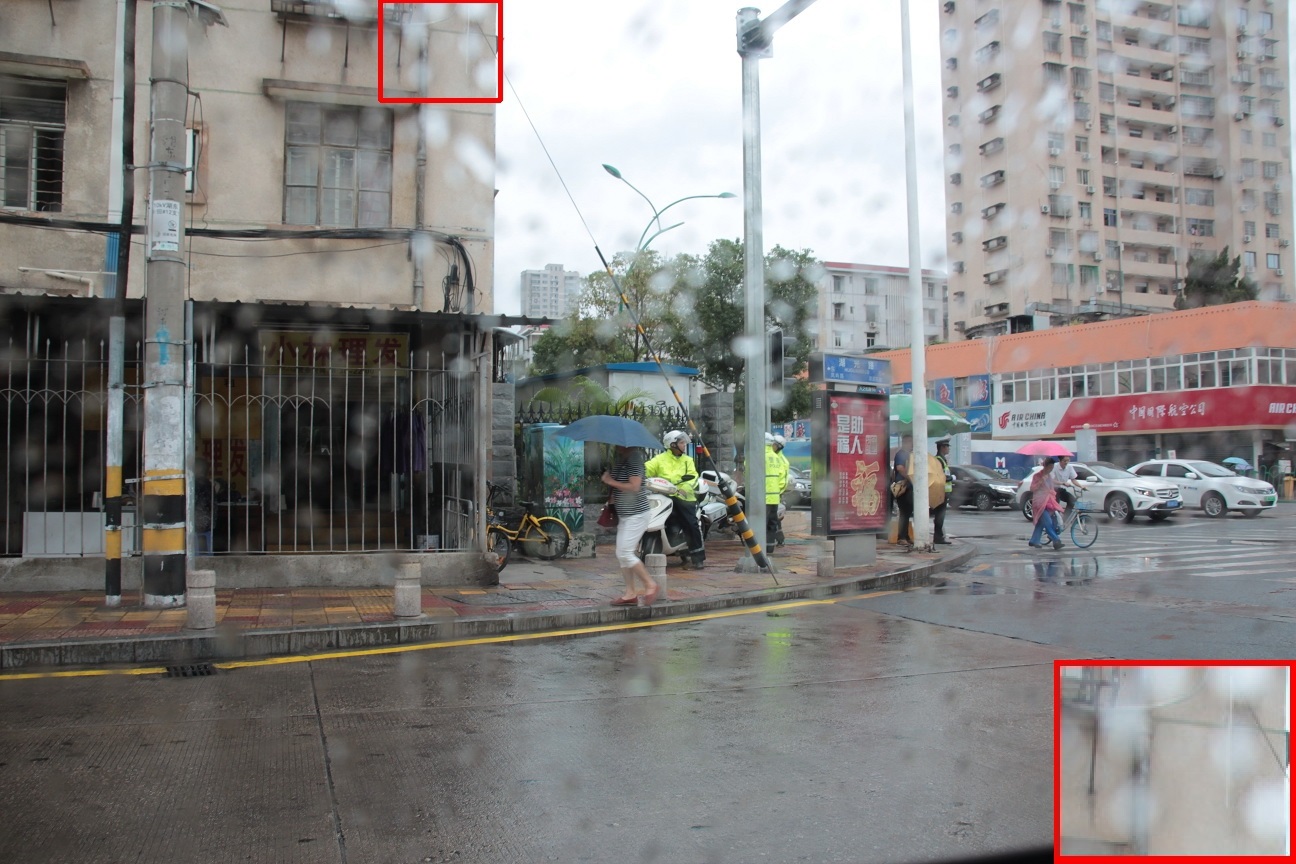}
	\includegraphics[width=1.2in]{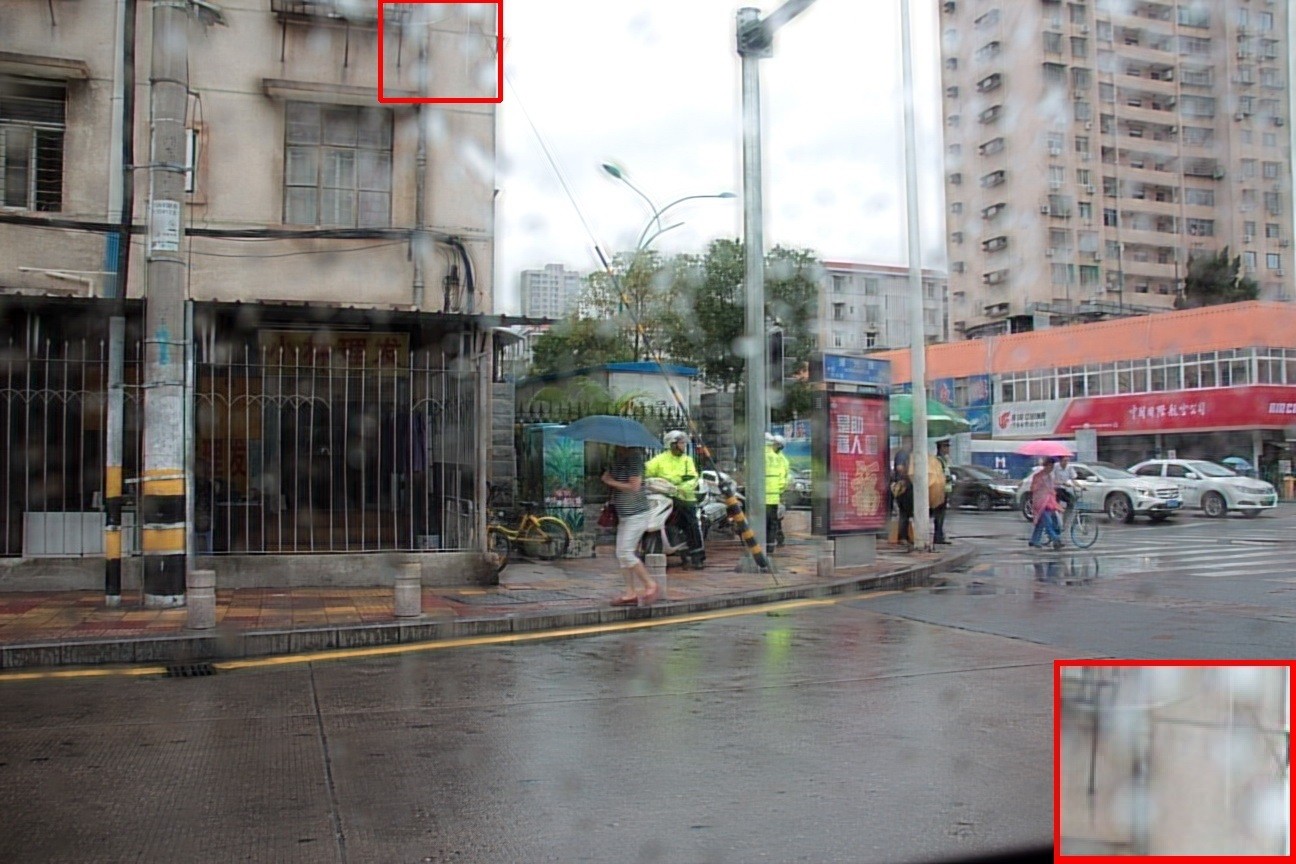}
	\includegraphics[width=1.2in]{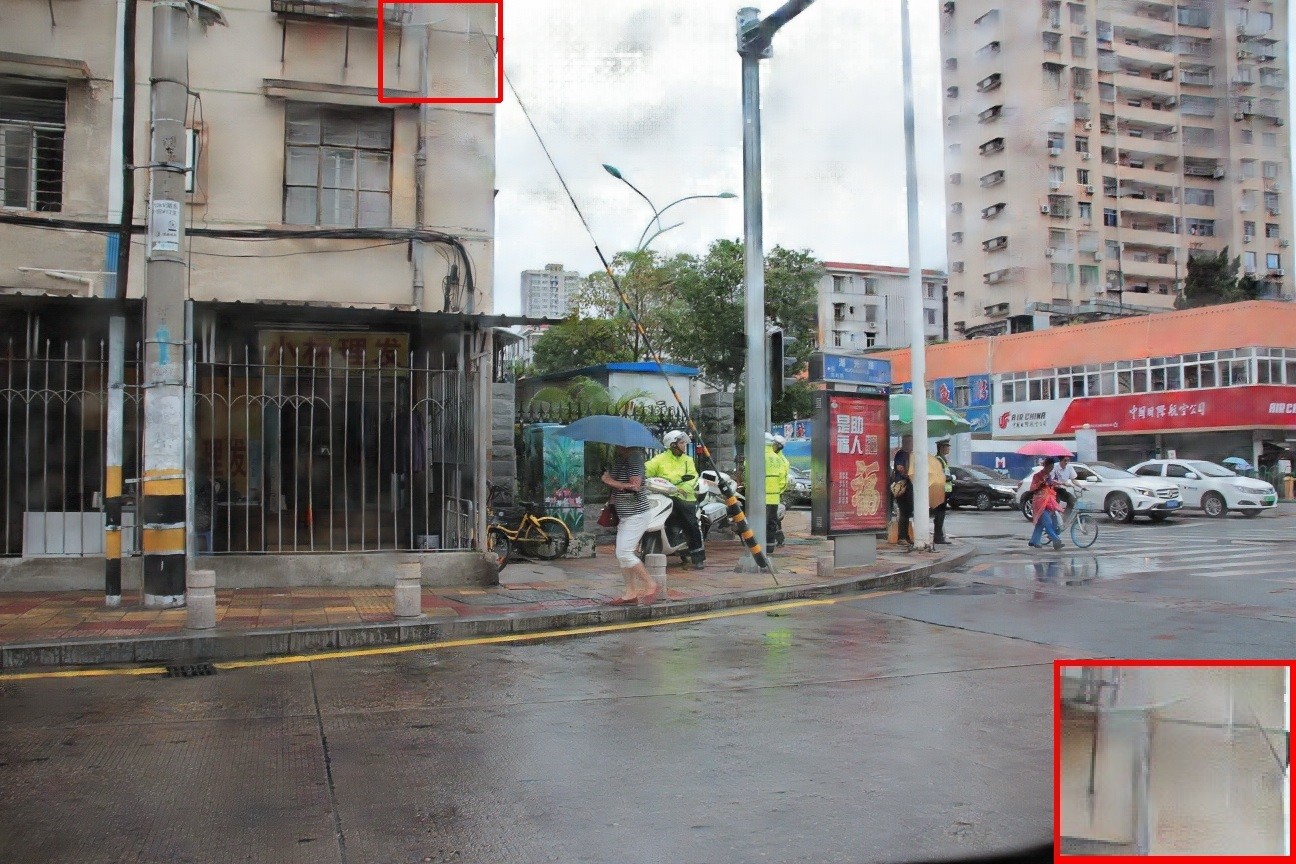}
	\includegraphics[width=1.2in]{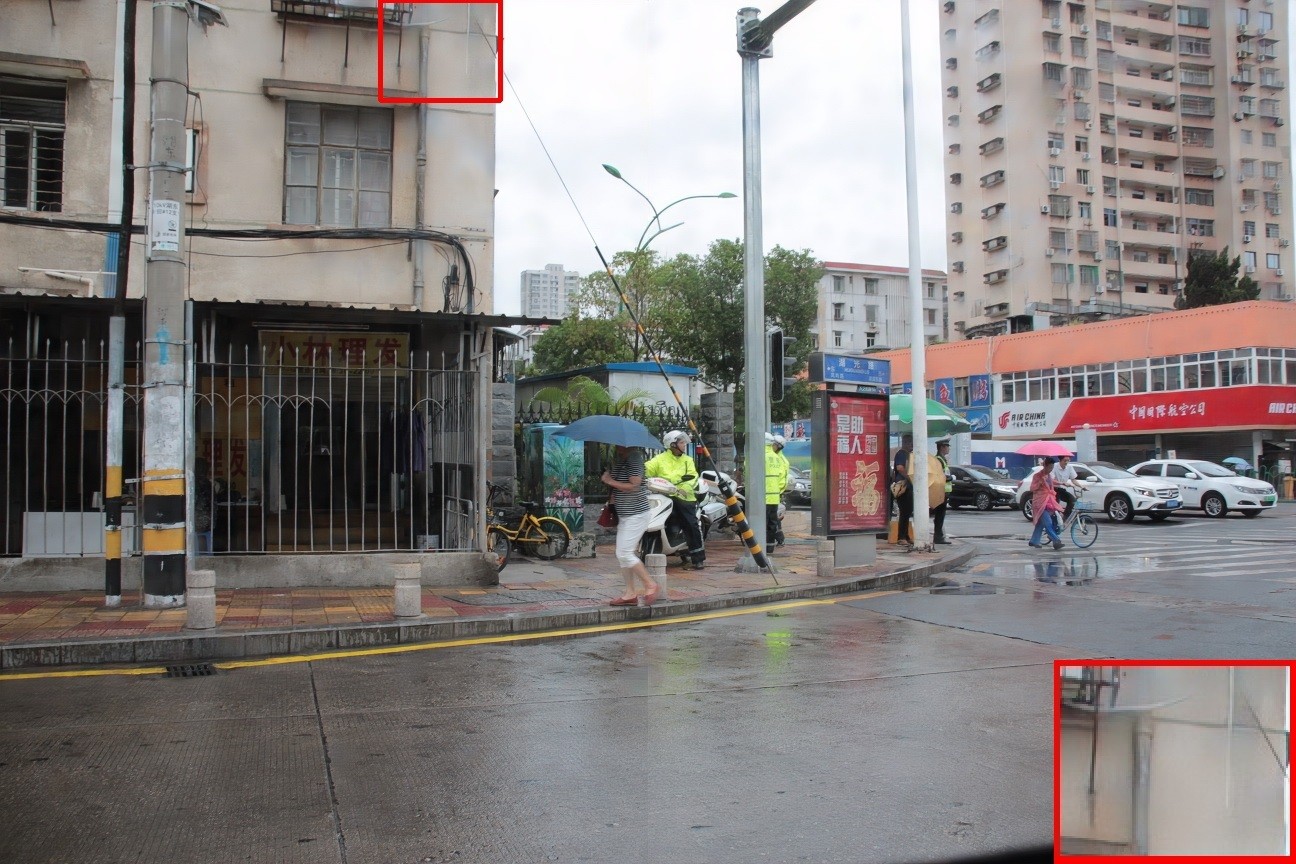}
	\includegraphics[width=1.2in]{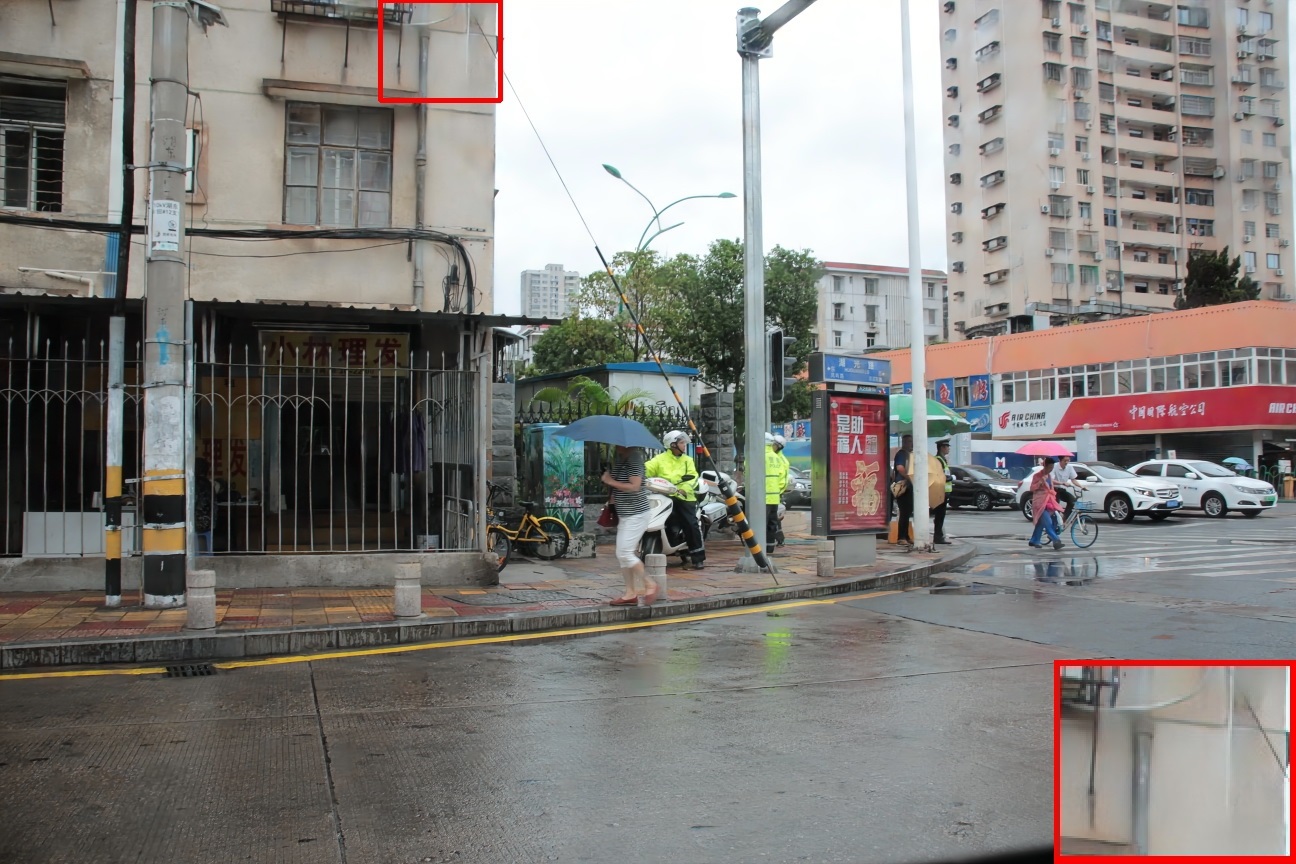}\\  \vspace{-0.065in}
	\subfigure[Input]{\includegraphics[width=1.2in]{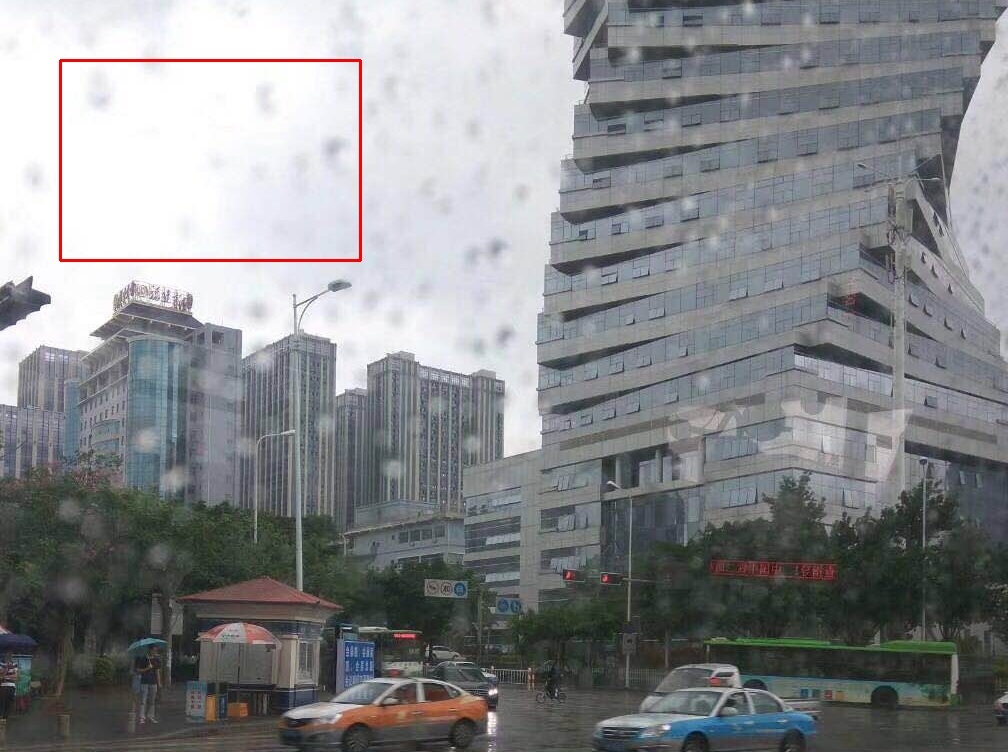}}
	\subfigure[Eigen \cite{eigen2013restoring}]{\includegraphics[width=1.2in]{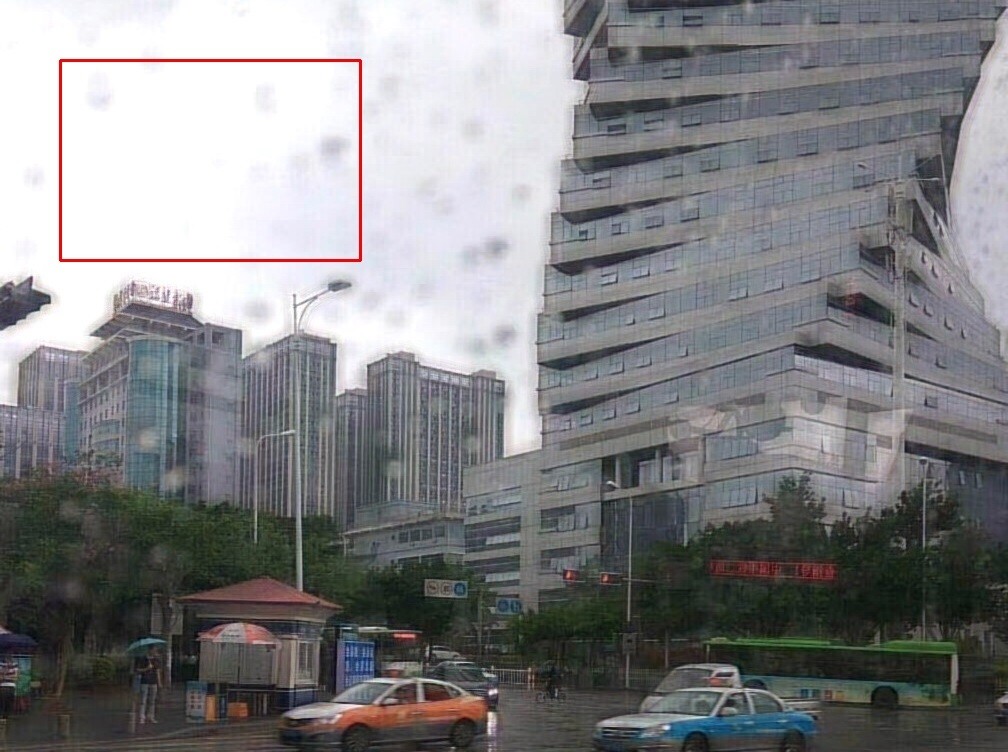}}
	\subfigure[Pix2pix \cite{isola2017image}]{\includegraphics[width=1.2in]{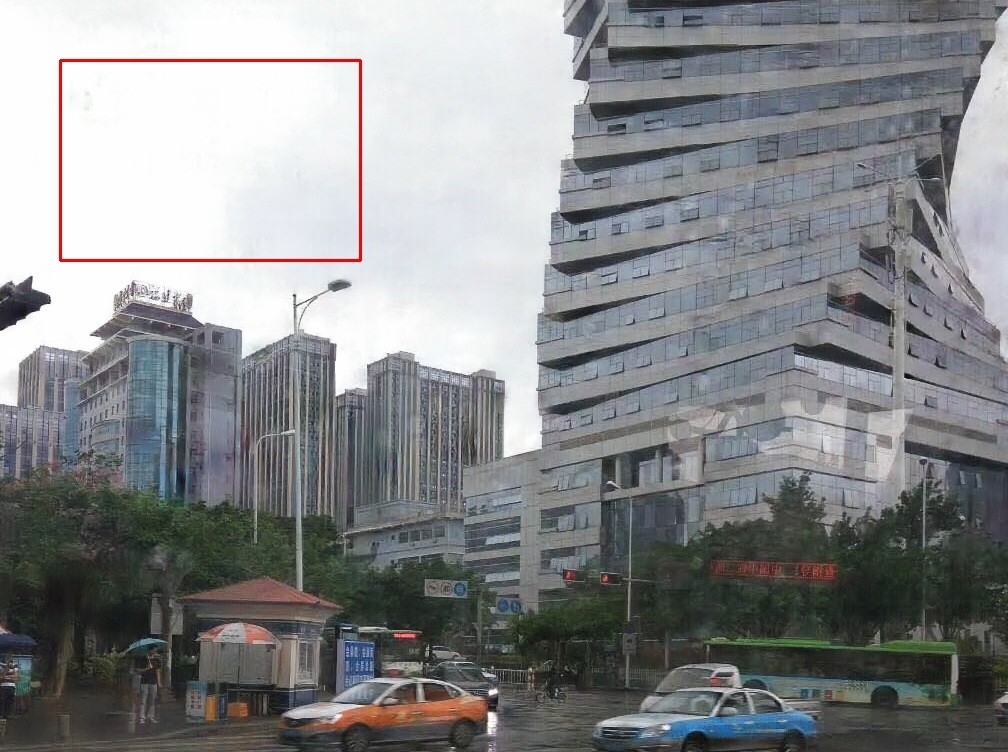}}
	\subfigure[AttentiveGAN \cite{qian2018attentive}]{\includegraphics[width=1.2in]{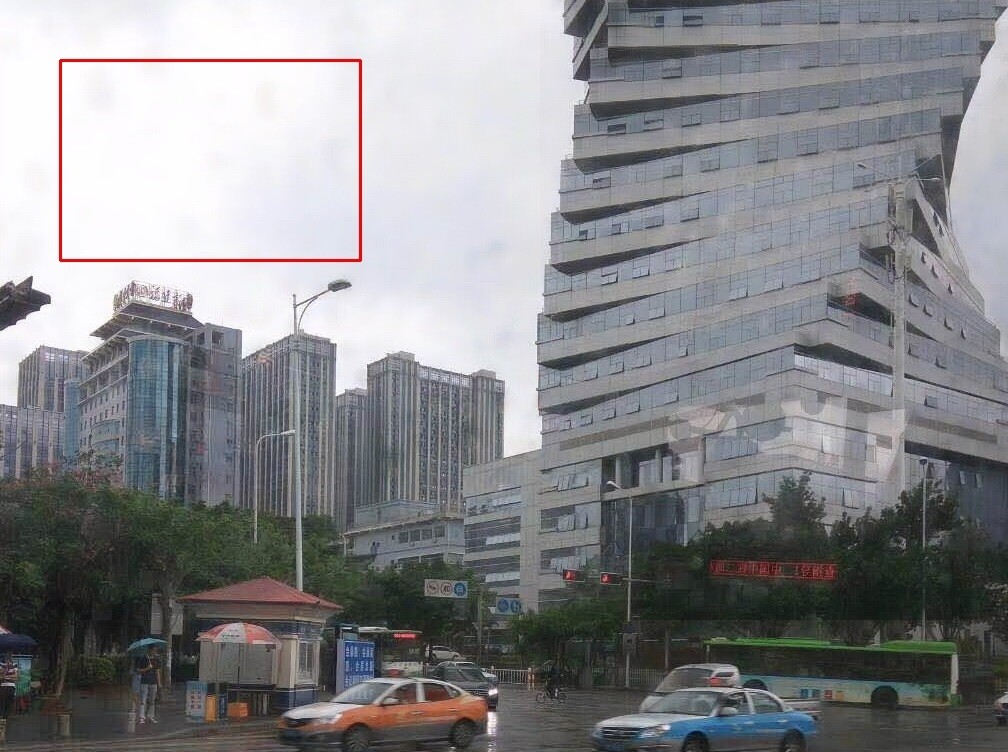}}
	\subfigure[Our A$^2$Net]{\includegraphics[width=1.2in]{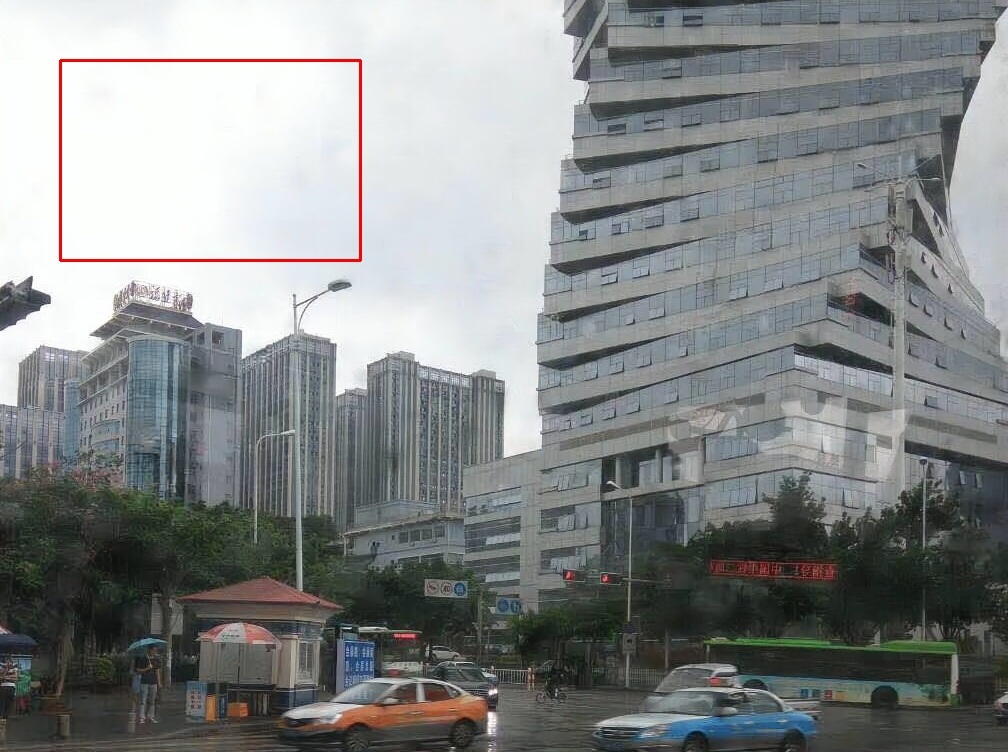}}
	\caption{Visual comparisons on real-world raindrop images.}
	\label{fig10}
\end{figure*}
\subsection{Training details}
We use the dataset from \cite{qian2018attentive}, which contains 861 pairs of clean/raindrop images, to train our networks. We randomly select 18,000 patch pairs with size of $256\times256$ in the training stage. We keep the learning rate to $2\times10^{-4}$ for the first 100 epochs and linearly decay the rate to 0 for the next 100 epochs. The mini-batch size is set to 4. All experiments are performed on a server with Inter(R) Core(TM) i7-8700K CPU and NVIDIA GeForce GTX 1080 Ti.

\section{Experiments}
We compare our A$^2$Net with three deep learning methods: Eigen \cite{eigen2013restoring}, Pix2pix \cite{isola2017image} and AttentiveGAN \cite{qian2018attentive}. For fair comparison, we use the raindrop dataset from \cite{qian2018attentive} to retrain all these methods.


\begin{table*}
	\caption{Comparison of running time (seconds).}
	\centering
	\begin{tabular}{|c|c|c|c|c|c|c|c|c|c|}
		\hline
		\multicolumn{2}{|c|}{} & \multicolumn{2}{|c|}{Eigen \cite{eigen2013restoring}} & \multicolumn{2}{|c|}{Pix2pix \cite{isola2017image}} & \multicolumn{2}{|c|}{AttentiveGAN \cite{qian2018attentive}} &  \multicolumn{2}{|c|}{Our  A$^2$Net}\\
		\hline
		\multicolumn{2}{|c|}{Image size} &CPU & GPU&CPU & GPU&\ \ CPU \ \ & GPU&CPU & GPU \\
		\hline
		\multicolumn{2}{|c|}{$256\times256$} &9.95 & 0.13&30.7 & 0.11&11.16 & 0.09&1.4 & 0.07 \\
		\hline
		\multicolumn{2}{|c|}{$512\times512$} &197.5 & 0.22&605.1 & 1.13&104.1 & 0.22&5.4 & 0.11 \\
		\hline
		\multicolumn{2}{|c|}{$640\times640$} &310.1 & 0.31&872.3 & 1.67&163.6 & 0.31&6.9 & 0.13 \\
		\hline
	\end{tabular}
	\label{table2}
\end{table*}

\begin{figure*}
	\centering
	\subfigure[Ground truth]{\includegraphics[width=1.1in]{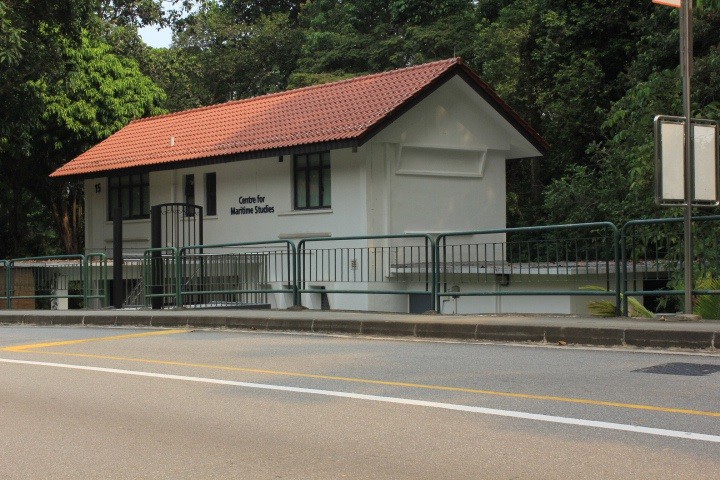}}
	\subfigure[Raindrop image]{\includegraphics[width=1.1in]{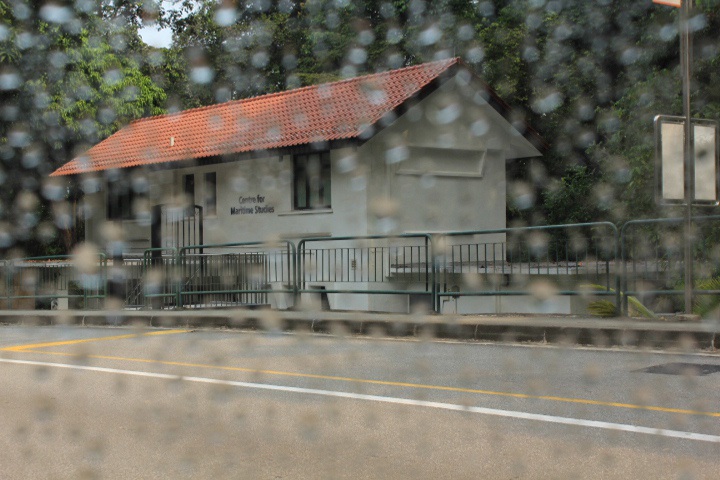}}
	\subfigure[generalNet]{\includegraphics[width=1.1in]{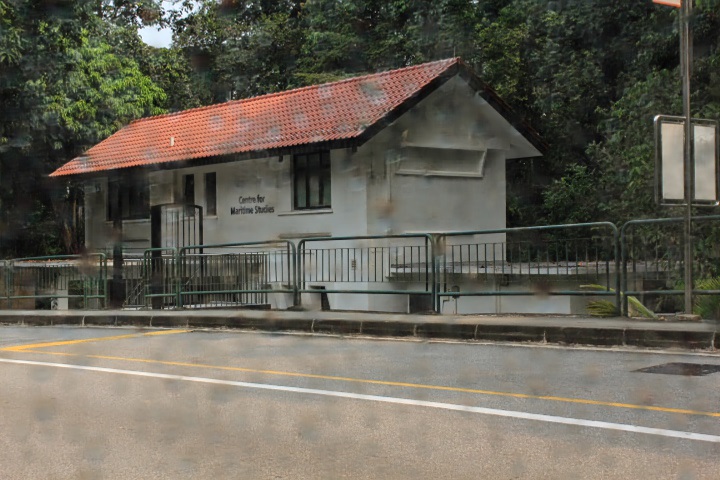}}
	\subfigure[A$^2$Net$^{\it{RGB}}$]{\includegraphics[width=1.1in]{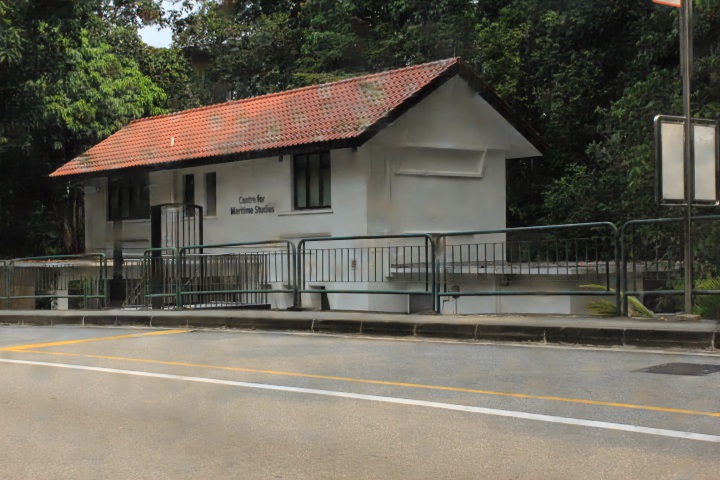}}
	\subfigure[A$^2$Net$^{\it{YUV}}$]{\includegraphics[width=1.1in]{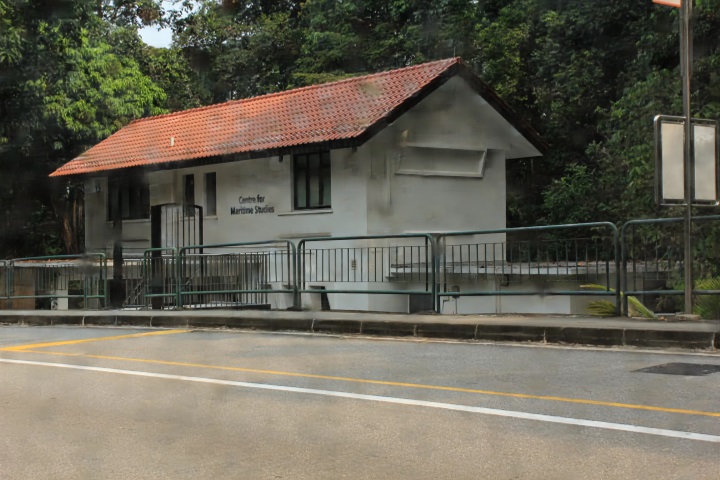}}
	\subfigure[A$^2$Net]{\includegraphics[width=1.1in]{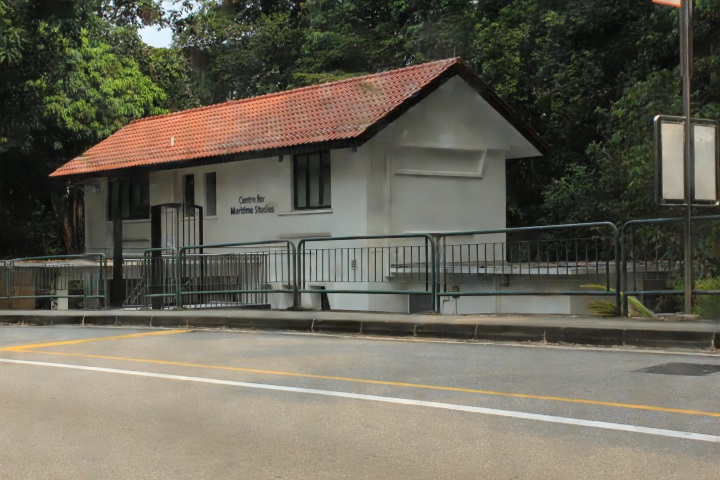}}
	\caption{Visual comparison on ablation study. }
	\label{fig11}
\end{figure*}

\subsection{Results on raindrop dataset}

There are two test dataset,  testA and testB, provided by \cite{qian2018attentive}. The testA, which contains 58 pairs of well aligned images, is a subset of testB.  The testB contains 249 pairs of images that better reflect the complexity of the raindrop distribution, but with a small portion of the image pairs that are not perfectly aligned. We test the above methods on both testA and testB to more accurately compare the performance of the above methods.

Figure \ref{fig9} shows visual comparisons for three raindrop-degraded images used for testing. As can be seen, Eigen \cite{eigen2013restoring} fails to remove raindrops due to the limitation of its model capacity. Pix2pix \cite{isola2017image} is able to remove the raindrops while tend to generate obvious artifacts. Our A$^2$Net has comparable visual results with the AttentiveGAN \cite{qian2018attentive} and outperforms other methods.

SSIM and PSNR are also adopted to perform quantitative evaluations in Table \ref{table1}. As shown in the Table \ref{table1}, our SSIM is higher than other methods on the testA. This indicates that our method can generate results more similar to the ground truth. Though our result has a lower PSNR value than AttentiveGAN method on testA, it does not affect the visual quality. This is because PSNR is calculated based on the mean squared error (MSE), which measures global pixel errors without considering local image characters. While testing on the testB, the performance of all methods has degraded. Imperfect alignment of a small number of image pairs may result in the decrease of SSIM and PSNR, but the main reason is that the distribution of raindrops in the testB is more complex which requires the methods to be more robust. As can be seen, our method achieves the highest SSIM and PSNR on testB. Moreover, our A$^2$Net contains the far fewer parameters, potentially making it more suitable for practical applications.

On the other hand, we tune the kernel numbers for all encoder and decoders to 32, and refer to this model as A$^2$Net-32. As shown in Table \ref{table11}, the performance of A$^2$Net-32 is slightly better than that of the final A$^2$Net, but the number of parameters is increased to 0.45M. This experiment further confirms that the UV channels does not have too much impact on the effect of raindrop removal. Therefore, we set the kernel numbers to 32, 32 and 24 as the default setting to balance the performance and parameters burden.
\phantom{anything}
\vspace{-0.2em}
\begin{table}[h]
	\caption{Comparison on increasing kernel numbers.}
	\centering
	\begin{tabular}{|c|c|c|c|c|c|c|c|c|c|c|}
		\hline
		& \multicolumn{2}{|c|}{Our A$^2$Net-32} & \multicolumn{2}{|c|}{Our A$^2$Net}\\
		\hline
		&SSIM  & PSNR &SSIM  & PSNR \\
		\hline
		{testA} &0.928 & 30.83&0.927 & 30.79 \\
		\hline
		{testB} &0.834 & 25.53&0.834 & 25.50 \\
		\hline
		{Parameters \#}  & \multicolumn{2}{|c|}{0.45M} & \multicolumn{2}{|c|}{0.40M}\\
		\hline
	\end{tabular}
	\label{table11}
\end{table}
\subsection{Results on real-world data}
In this section, we show that A$^2$Net, which is trained on the raindrop dataset \cite{qian2018attentive}, can still works well on real-world data. We collect 100 real-world raindrop images from the Internet as a new dataset\footnote {Our code and data will be released soon.}. Since no ground truth exists, we only show the qualitative results on real-world data in Figure \ref{fig10}. As can be seen, our A$^2$Net achieves the best visual performance on removing raindrops and preserving details compared to other methods.


\subsection{Running time}
Our A$^2$Net can process new images very efficiently. Table \ref{table2} shows the average running time of 100 test images for three different image sizes. Our method has the faster computational time on both CPU and GPU compared with other deep models\footnote {Please see the supplement for more comparisons.}. As a pre-processing for other high-level vision tasks, the raindrop removal process should be simpler and faster. However, the running time of Eigen, Pix2pix and AttentiveGAN on the CPU increases rapidly with the increase of image size, the computing resources required are enormous. Our A$^2$Net is a relatively shallow network that requires fewer calculations, so it is more practical, for example, on mobile devices.

\subsection{Ablation study}
To validate the necessity of color space conversion strategy and adjacent feature aggregation operations, we design three variants of the A$^2$Net for exhaustive comparison. One is called A$^2$Net$^{\it{YUV}}$ that only keeps one aggregation encoder and one aggregation decoder. The A$^2$Net$^{\it{YUV}}$ directly learns the image-to-image mapping by focusing on all YUV channels in YUV space. The second is called A$^2$Net$^{\it{RGB}}$ which has the same structure as A$^2$Net$^{\it{YUV}}$, but is trained in RGB space. The last one is called generalNet that only keeps one general encoder and one general decoder, as shown in Figure \ref{A2network}(a). The generalNet is also trained in RGB space. We adopt global residual learning for each network structure.

\begin{figure}
	\centering
	\subfigure[Comparison of SSIM using testA]{\includegraphics[width=0.72\linewidth,trim=0 5 0 0,clip]{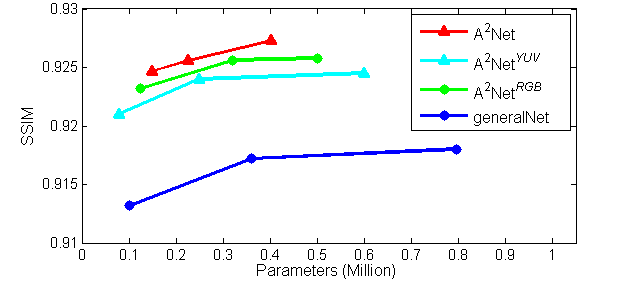}}\\
	\subfigure[Comparison of SSIM using testB]{\includegraphics[width=0.72\linewidth,trim=0 5 0 0,clip]{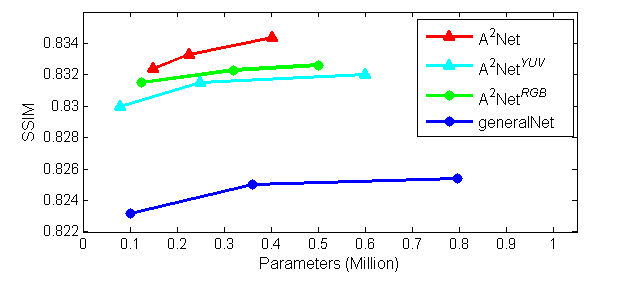}}
	\caption{Comparison of using different color space.}
	\label{fig12}
\end{figure}

We show the comparison of the above networks in Figure \ref{fig12}. Since these networks have similar numbers of layers, we only tune their kernel numbers to control the parameters. We find that A$^2$Net$^{\it{RGB}}$ can achieve higher SSIM and better parameters efficiency than generalNet. This means that the adjacent aggregation operation can effectively enhance the network capacity. On the other hand, the A$^2$Net$^{\it{YUV}}$ cannot produce satisfactory results since it does not handle the Y and UV channels separately. Our final A$^2$Net achieves the best raindrop removal performance. This is because the raindrop removal problem can be simplified with the color space conversion strategy.

Figure \ref{fig11} shows one visual result. As can be seen in Figure \ref{fig11}(c), most of the raindrops in the image can be removed via the generalNet, but the quality of the reconstructed image is not desirable. The adjacent aggregation operation helps a lot to preserve image details, and the color space conversion strategy can further improve the raindrop removal performance, as shown in Figures \ref{fig11}(d) and (f).
\subsection{Loss Function}
\begin{figure}
	\centering
	\subfigure[Raindrop image]{\includegraphics[width=1.5in]{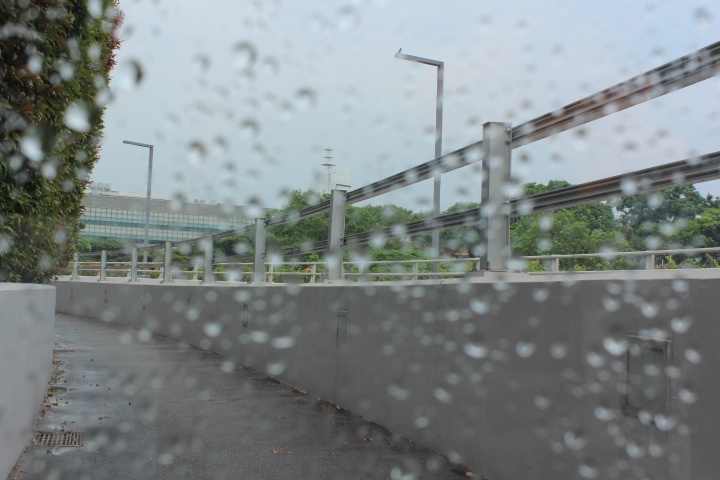}}
	\subfigure[Using only MSE loss]{\includegraphics[width=1.5in]{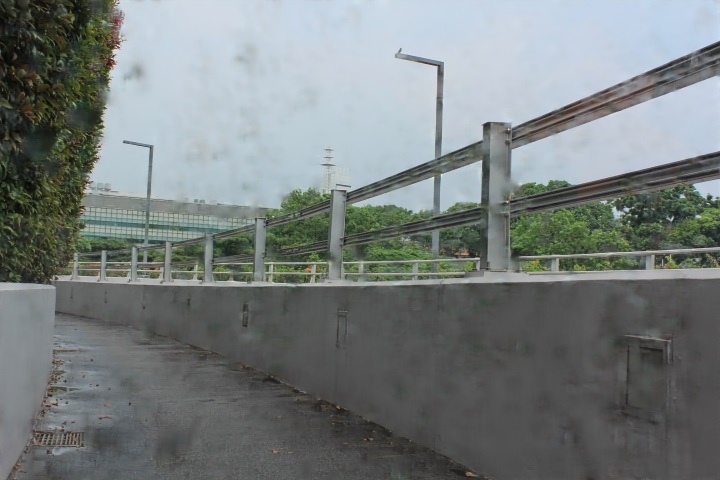}}\\
	\subfigure[Using only SSIM loss]{\includegraphics[width=1.5in]{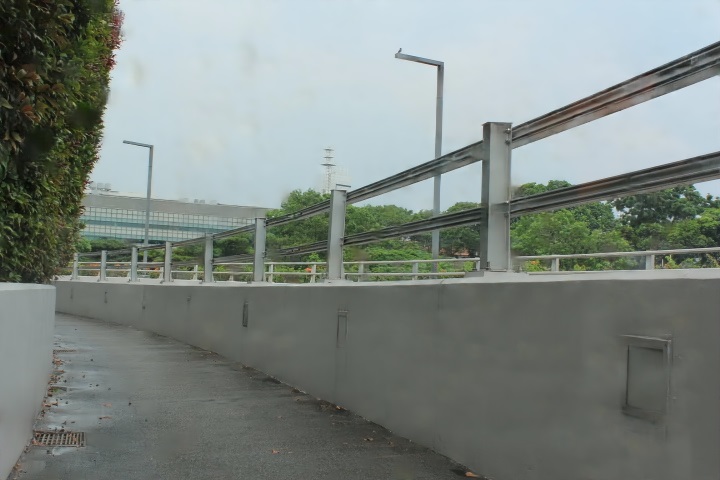}}
	\subfigure[Using MSE + SSIM loss]{\includegraphics[width=1.5in]{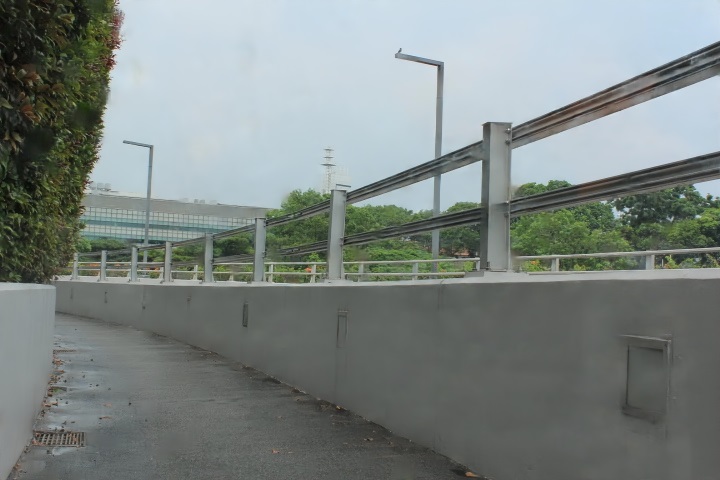}}
	\caption{An example by using different losses.}
	\label{fig13}
\end{figure}

In this section, we discuss different loss functions to study their impact on performance. Figure \ref{fig13} shows one visual comparison  by using different loss functions. As can be seen, using only MSE loss generates overly smooth results, while the SSIM loss can preserve details well. Moreover, combining the SSIM and MSE losses can generate images with better global structure and clearer details, as shown in Figure \ref{fig13}(d). Table \ref{table3} also demonstrates that using the combined loss (\ref{eq.6}) can further improve the performance and produce visual pleasing results.

\begin{table}
	\caption{Comparison on using different loss functions.}
	\small
	\centering
	\begin{tabular}{|c|c|c|c|c|c|c|}
		
		\hline
		{Loss} & \multicolumn{2}{|c|}{MSE} & \multicolumn{2}{|c|}{SSIM} & \multicolumn{2}{|c|}{Eq. (\ref{eq.6})} \\
		
		\hline
		
		{} &SSIM &PSNR&SSIM &PSNR&SSIM &PSNR \\
		\hline
		testA &0.861 &26.48&0.926 &30.59&0.927 &30.79 \\
		\hline
		testB &0.785 &24.07&0.834 &25.33&0.834 &25.50 \\
		\hline
	\end{tabular}
	\label{table3}
\end{table}

\section{Conclusion}
In this paper, we have proposed a adjacent aggregation network for single image raindrop removal. Our network employs an encoder-decoder structure with a new adjacent aggregation operation to effectively extract informative features. To simplify the raindrop removal problem, our A$^2$Net learns the raindrop removal function in YUV color space. By this way, the complex image-to-image mapping problem can be transformed into luminance (Y channel) and chrominance (UV channels) processing problems. Most of the network's attention is focused on the Y channel. Our A$^2$Net contains only 0.4M parameters while can still achieve state-of-the-art performance.

{\small
\bibliographystyle{ieee}
\bibliography{egbib}
}

\end{document}